\documentclass[letterpaper]{article} 
\usepackage[preprint]{aaai2027}  
\usepackage[hyphens]{url}  
\usepackage{graphicx} 
\usepackage{natbib}  
\usepackage{caption} 
\usepackage{algorithm}
\usepackage{algorithmic}
\usepackage{longtable}
\usepackage{newfloat}
\usepackage{listings}
\DeclareCaptionStyle{ruled}{labelfont=normalfont,labelsep=colon,strut=off} 
\floatstyle{ruled}
\newfloat{listing}{tb}{lst}{}
\floatname{listing}{Listing}

\usepackage{booktabs}
\usepackage{multirow} 
\usepackage{amssymb}
\usepackage{bbding}
\usepackage{pifont}
\usepackage{xspace}
\usepackage{amsmath}
\usepackage[table,xcdraw]{xcolor}
\usepackage{booktabs}

\newcommand{\paragraphb}[1]{\noindent\textbf{#1}}
\newcommand{\name}{\texttt{RoboGraph}\xspace}
\newcommand{\benchname}{\texttt{RoboGraphBench}\xspace}

\title{Compiling and Benchmarking Task-State Horizons for \\ Embodied Agents}
\author{
    Meiqi Wang\textsuperscript{\rm 1}\equalcontrib,
    Shichao Li\textsuperscript{\rm 2}\equalcontrib
}
\affiliations{
    \textsuperscript{\rm 1}Tsinghua University\\
    \textsuperscript{\rm 2}Beijing University of Posts and Telecommunications\\
    wang-mq22@mails.tsinghua.edu.cn, lishichao@bupt.edu.cn
}

\begin{document}

\maketitle

\begin{abstract}
Frontier agentic models are increasingly deployed as high-level planners for long-horizon embodied tasks.
Existing robotic benchmarks have advanced long-horizon evaluation, but primarily characterize difficulty through action-sequence length and subtask complexity, overlooking a distinct challenge: agents must track evolving task-relevant world states induced by both their exploration and environmental dynamics. 
We define the span of task-relevant state transitions that an agent must track as task-state horizon (TSH).
To evaluate how agent performance varies with TSH, we introduce \name, a robotic task compiler that translates state-transition dependencies into executable symbolic graphs.
Specifically, \name constructs task-state horizons from spatial and temporal causal dependencies, including those induced by unexpected failures and interventions during task execution.
Building on~\name, we release a benchmark comprising 588 episodes across 84 scenes with varying TSHs.
Experiments evaluating {15} advanced agentic models in both semantic and visual closed-loop environments show that most models struggle with demanding TSHs, revealing substantial gaps in maintaining, exploring, and updating task-relevant state over long horizons.

\end{abstract}

\section{Introduction}

Recent advances in embodied foundation models have produced agents capable of interpreting language, grounding visual observations, and generating action plans~\cite{ahn2022can,intelligence2025pi_,intelligence2604pi0,wu2026pragmatic,wang2026qwen}.
{To assess long-horizon action planning,} existing benchmarks make substantial progress in {task composition~\cite{shridhar2020alfred,li2024behavior,chang2025partnr}, embodied memory~\cite{yadav2026findingdory,hu20263dllm}, and adaptation to changing environments~\cite{liao2026spamem,zhang2026worldlines}}, for instance, to traverse more rooms, manipulate more objects, and satisfy more subgoals.
{However, action and subgoal complexity capture only part of embodied task difficulty.}
\textit{A further challenge lies in managing evolving task-relevant world state over extended horizons.}
For instance, a decision at step $t$ may depend on state information obtained at step $t-k$, which was initially hidden and required active exploration, or unexpectedly changed in the interim and required world state updating.

\begin{figure}[t]
    \centering
    \includegraphics[width=0.9\linewidth]{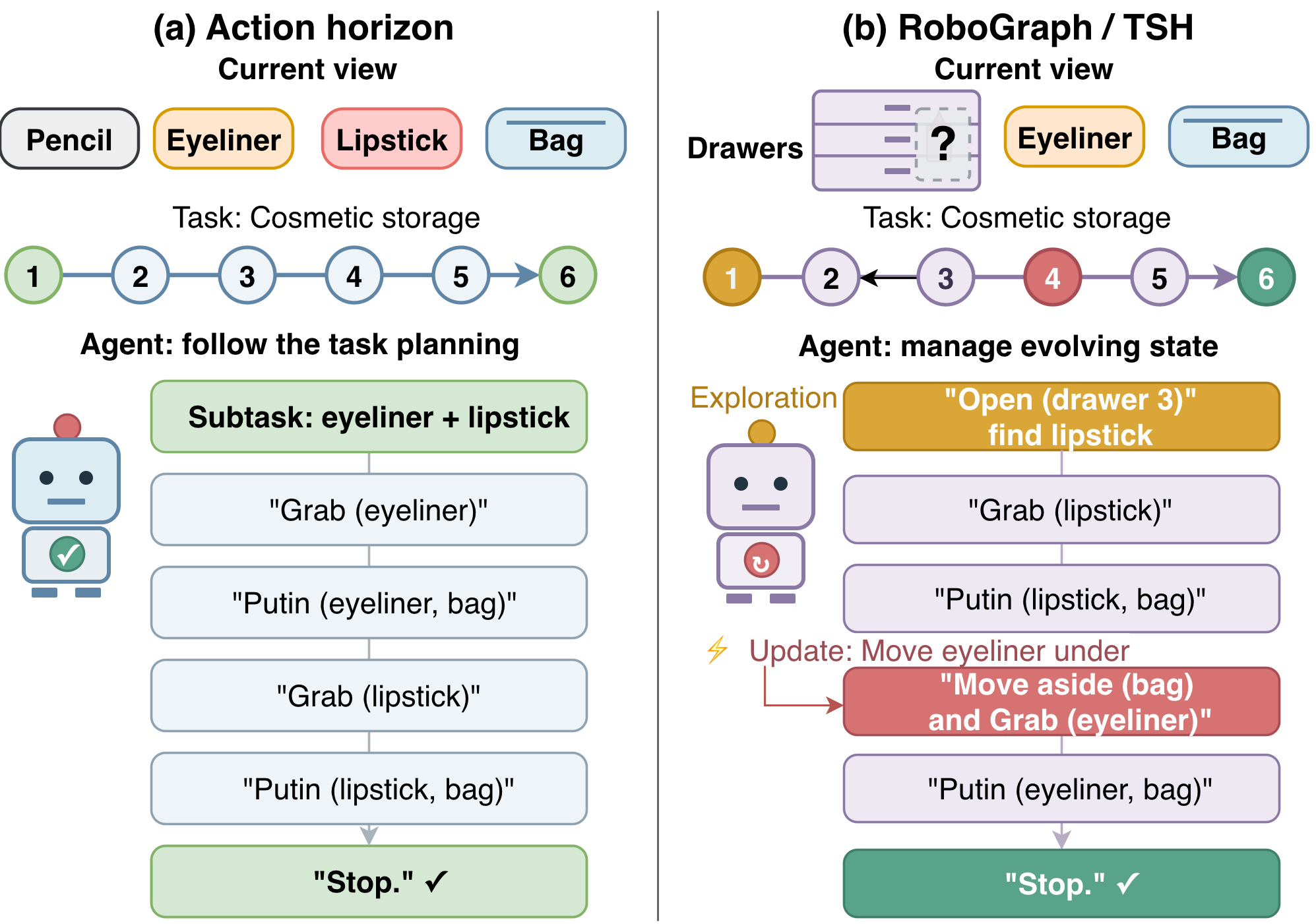}
    \vspace{-1ex}
    \caption{An example illustrating the distinction between the proposed task-state horizon (TSH) and action horizon.}
    \label{fig:examples}
    \vspace{-2ex}
\end{figure}

Crucially, long-sequence actions and complex world states are not always aligned.
As illustrated in \figurename~\ref{fig:examples}, both (a) and (b) require an agent to place a lipstick into a storage bag. In (b), however, the lipstick is out of sight: it may occupy one of several drawers or be moved during execution. 
Thus, despite an identical action sequence, (b) imposes more demanding state-transition dependencies.

We formalize this challenge through the \textit{task-state horizon} (TSH), which characterizes \textit{the span over which an agent must maintain, explore, and update task-relevant world state} to sustain reliable longer-horizon execution.
Specifically, we decompose state-management demands into three dimensions: \textit{state maintenance}, retaining and retrieving previously observed state; \textit{state exploration}, actively resolving hidden or uncertain state; and \textit{state updating}, detecting environmental changes and revising outdated beliefs.
Whereas most existing embodied benchmarks implicitly entangle task-relevant state dependencies within subtasks~\cite{shridhar2020alfred,li2024behavior,zhang2025plan,zhu2026robots}, TSH formulation makes both state dependencies and the associated transitions explicit.
Furthermore, this enables fine-grained attribution of unsuccessful subtasks to specific state-management errors; for instance, forgetting the lipstick’s location indicates a maintenance failure, neglecting to inspect candidate drawers indicates an exploration failure, and relying on outdated location after moving indicates an updating failure.

To enable systematic measurement and controlled variation of TSH, we introduce \name, a robotic task compiler that translates spatial and temporal causal dependencies among task-relevant state transitions into executable symbolic graphs.
Spatial dependencies capture state propagation across objects, while temporal dependencies capture how prior states constrain later actions.
Each graph explicitly represents these dependencies and records transitions induced by agent actions or environmental changes. 
The graph serves as an intermediate representation that can be executed across multiple backends, including symbolic environments, simulators, and real-world robots.

Given a task family goal, we automatically generate numerous graphs embedding varying spatial and temporal dependencies.
We construct \benchname, a benchmark comprising 10 task families including tabletop manipulation and indoor navigation, with 84 automatically generated scenes.
Furthermore, we incorporate six types of external interventions during execution, thus yielding a total of 588 episodes.
We evaluate {15} agentic models across closed- and open-source families using our unified harness in both semantic and visual closed-loop environments.
Beyond task success rate, we introduce new metrics to assess agents' state management capabilities.
Although GPT-5.5 performs comparatively well, it still exhibits imperfect state maintenance through redundant actions, insufficient exploration through premature termination, and limited state updating through prolonged recovery latency.
These results reveal substantial remaining gaps in handling tasks with demanding TSHs.

In conclusion, we make three contributions:
\begin{itemize}
    \item We introduce the {task-state horizon} (TSH) as a complement to existing action-horizon measures, capturing the task-relevant state-management demands of robotic agents, i.e., maintenance, exploration, and updating.
    \item We design \name, a compiler that transforms spatial and temporal state dependencies into executable semantic scene graphs and supports dynamic world changes to control varying TSHs.
     \item We introduce \benchname, instantiating 588 TSH-controlled robotic tasks, new metrics to systematically assess state management, and a unified harness to evaluate 15 frontier agentic models in both symbolic and simulated closed-loop environments.
\end{itemize}

\section{Related Work}

\paragraphb{Long-horizon embodied tasks.}
Embodied benchmarks have progressively expanded the compositional,
spatial, and temporal structure of goal-directed interaction.
VirtualHome \cite{puig2018virtualhome}, ALFRED \cite{shridhar2020alfred}, and ALFWorld \cite{shridhar2020alfworld} study programmatic or language-grounded task execution in household environments.
BEHAVIOR \cite{srivastava2022behavior} and BEHAVIOR-1K \cite{li2024behavior} scale this setting to large collections of predicate-defined activities with richer object states, action preconditions, and physical interactions.
PARTNR \cite{chang2025partnr}, ET-Plan-Bench \cite{zhang2025plan}, and LongAct \cite{zhu2026robots} further extend long-horizon evaluation to temporal dependencies, coordination constraints, and interdependent household goals.
Across these benchmarks, long-horizon difficulty is largely expressed through the scale and structure of actions, subgoals, and constraints.
\name{} extends this view by identifying state management as a core challenge in long-horizon embodied tasks.

\paragraphb{Long-term memory and changing world state.}
Long-horizon action planning requires agents to use information accumulated over extended interactions.
FindingDory \cite{yadav2026findingdory}, LMEE-Bench \cite{wang2026explore}, and 3DMem-Bench \cite{hu20263dllm} evaluate long-term memory in embodied navigation, exploration, and spatial reasoning.
MIKASA-Robo \cite{cherepanov2026memory}, RoboMME \cite{dai2026robomme}, and RoboMemArena \cite{robomemarena2025} focus on manipulation tasks that depend on prior states, actions, and event sequences.
SpaMEM \cite{liao2026spamem} and WorldLines \cite{zhang2026worldlines} study how agents track and revise world state as objects and relations evolve.
In our work, we identify the broader challenge underlying these capabilities: actively reasoning over accumulated and evolving state to guide task decisions throughout execution.

\paragraphb{Procedural task generation and benchmark compilation.}
Procedural generation has substantially expanded the scale of embodied environments and tasks. ProcTHOR \cite{procthor} generates diverse interactive homes, while GenSim \cite{wang2024gensim} and RoboGen \cite{wang2024robogen} generate simulation tasks, scenes, demonstrations, and training supervision.
PARTNR \cite{chang2025partnr} and ManiTaskGen \cite{dai2025manitaskgen} further incorporate simulator grounding and feasibility constraints into task generation.
 As embodied agents move toward longer-horizon and more autonomous tasks, \name{} compiles persistent and evolving state dependencies into executable task families.

\begin{figure*}[h]
    \centering
    \includegraphics[width=0.9\linewidth]{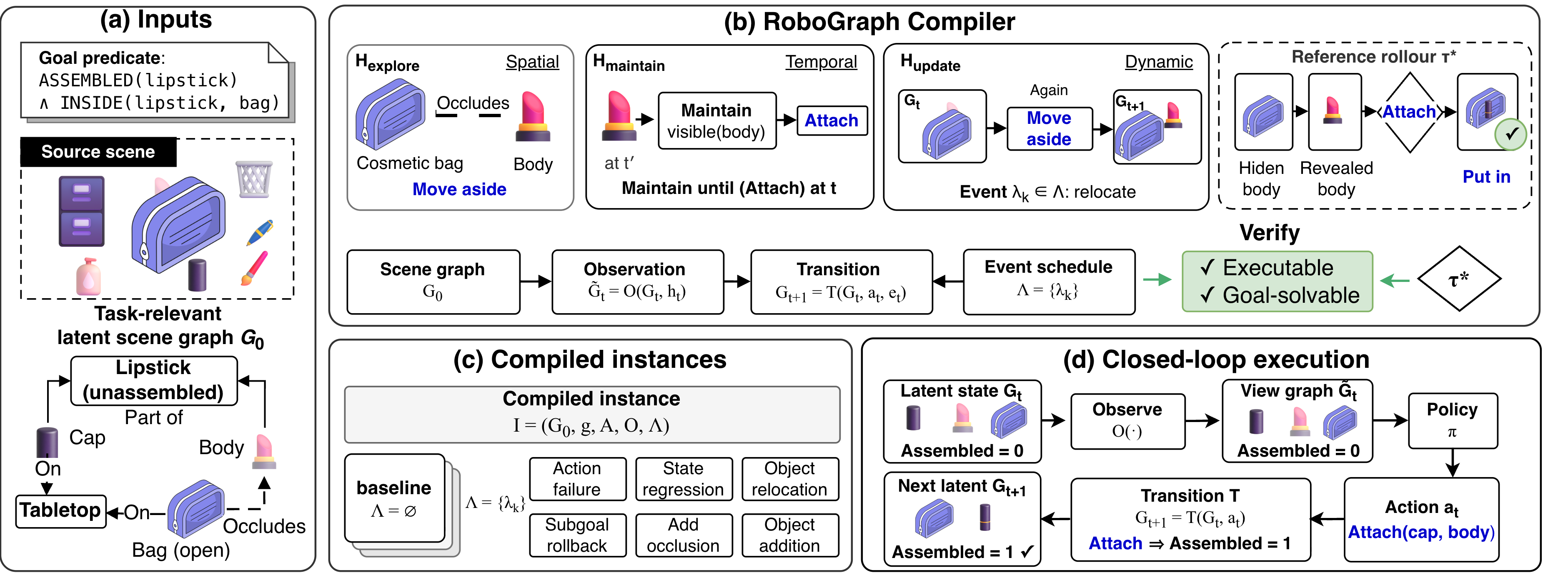} 
    \vspace{-1ex}
    \caption{
    {Overview of \name.}
    (a) Compiler inputs include a task-relevant latent scene graph and a goal predicate.
    (b) \name encodes spatial, temporal, and dynamic state dependencies into executable symbolic task graphs with controlled TSHs, verified through a reference rollout.
    (c) The compiler generates a baseline instance and intervention-augmented variants.
    (d) During closed-loop execution, the agent observes the partial view graph, selects an action, and updates the latent world state.
    }
   \label{fig:overview}
   \vspace{-2ex}
\end{figure*}

\section{Problem Formulation}
\label{sec:problem}
  \paragraphb{Embodied tasks under partial observation.}
We formulate embodied interaction as a partially observable decision process~\cite{kaelbling1998planning}.
  Let the task-relevant world state at time $t$ be a scene graph $G_t=(V_t,E_t,X_t)$, where nodes $V_t$ represent entities, edges $E_t$ represent spatial and functional relations, and $X_t$ contains node attributes.
  We denote that a \emph{fact} is an atomic assertion over $G_t$, e.g., an open drawer, and a base task is
  \begin{equation}
  \mathcal{B}=(G_0,g,\mathcal{A}),
  \end{equation}
  where $G_0$ is the initial state, $g$ is a goal predicate over graphs, and $\mathcal{A}$ is a set of task-level action schemas, each with
  state-dependent preconditions and effects, $a=\langle\mathrm{Pre}(a),\mathrm{Eff}(a)\rangle$.

  The agent never receives the complete graph. At each step, the environment returns a task-conditioned \emph{view graph}
  \begin{equation}
  \widetilde{G}_t=O(G_t,h_t),
  \label{eq:view}
  \end{equation}
  where $h_t$ is the interaction history and $O$ exposes the facts currently observable given the agent's past actions.
  A policy selects actions conditioned on the goal, the current view graph, and its action history,
  \begin{equation}
  a_t\sim\pi(\cdot\mid g,\widetilde{G}_{t},a_{0:t-1}).
  \end{equation}

Let $\mathcal{E}$ denote the set of admissible exogenous events,
including environment-induced state changes and execution disturbances.  
At each step, the world evolves according to both the agent-selected action and an optional event:
\begin{equation}
G_{t+1}=\mathcal{T}(G_t,a_t,e_t),
\qquad
e_t\in\mathcal{E}\cup\{\varnothing\},
\label{eq:trans}
\end{equation}
where $e_t$ denotes an environment-induced transition, e.g., an object being relocated, with $e_t=\varnothing$ when
no such transition occurs.
Task-relevant facts may therefore change without being reflected in $\widetilde{G}_t$ until the agent re-observes the affected region.
A rollout $\tau$ succeeds if the agent issues \texttt{stop} in a goal-satisfying state within the step budget:
\begin{equation}
\operatorname{Succ}(\tau)
=
\mathbb{I}
\left[\,
T\leq T_{\max},\;
a_T=\textsc{stop},\;
G_T\models g
\,\right].
\end{equation}
where $T$ denotes the step at which the rollout terminates.

\paragraph{Definition 1 (state dependency).}
A decision at step $t$ has a \emph{state dependency} on a fact $f$ if actions at $t$ depend on $f$, while $f$ is not recoverable from the current view graph $\widetilde{G}_t$ alone. 
State dependencies arise from two sources:
\begin{itemize}
    \item A \emph{spatial dependency} occurs when $f$ is hidden in the current state, e.g., behind occluders or inside containers, and must be
    exposed through view-exploring actions.
    \item A \emph{temporal dependency} arises when a fact $f$, observed at an
    earlier step $t'<t$, remains necessary for selecting a goal-directed
    action at step $t$. 
\end{itemize}
If an exogenous event changes $f$ during the interval $(t',t)$, the agent's retained estimate of $f$ becomes stale and must be updated before it can support the decision at $t$.

\paragraph{Definition 2 (Task-state horizon).}
Let $\tau^{*}$ be a \emph{reference solution} of a task instance $\mathcal{I}$, i.e., a shortest oracle action sequence, and let $\mathcal{F}$ be the set of facts on which some action decision along $\tau^{*}$ has a state dependency.
The \emph{task-state horizon} (TSH) of $\mathcal{I}$ is a tuple:
  \begin{equation}
  H=(H_{\mathrm{maintain}},\,H_{\mathrm{explore}},\,H_{\mathrm{update}}),
  \end{equation}
  where
  \begin{itemize}
      \item $H_{\mathrm{maintain}}$ is the maximum, over all temporally-dependent facts in $\mathcal{F}$, of the number of steps in $\tau^{*}$ between the acquisition of the fact and the last decision it constrains.
      \item $H_{\mathrm{explore}}$ is the maximum, over all spatially-dependent facts in $\mathcal{F}$, of the minimal number of view-expanding actions required to expose the fact in $G_0$.
      \item $H_{\mathrm{update}}$ is the number of facts in $\mathcal{F}$ invalidated by exogenous events along $\tau^{*}$.
  \end{itemize}
  As all components are computed on $\tau^{*}$, TSH is a property of the task instance itself and measures the state-management burden of efficient execution.

\begin{figure*}[t]
\centering
    \includegraphics[width=0.9\linewidth]{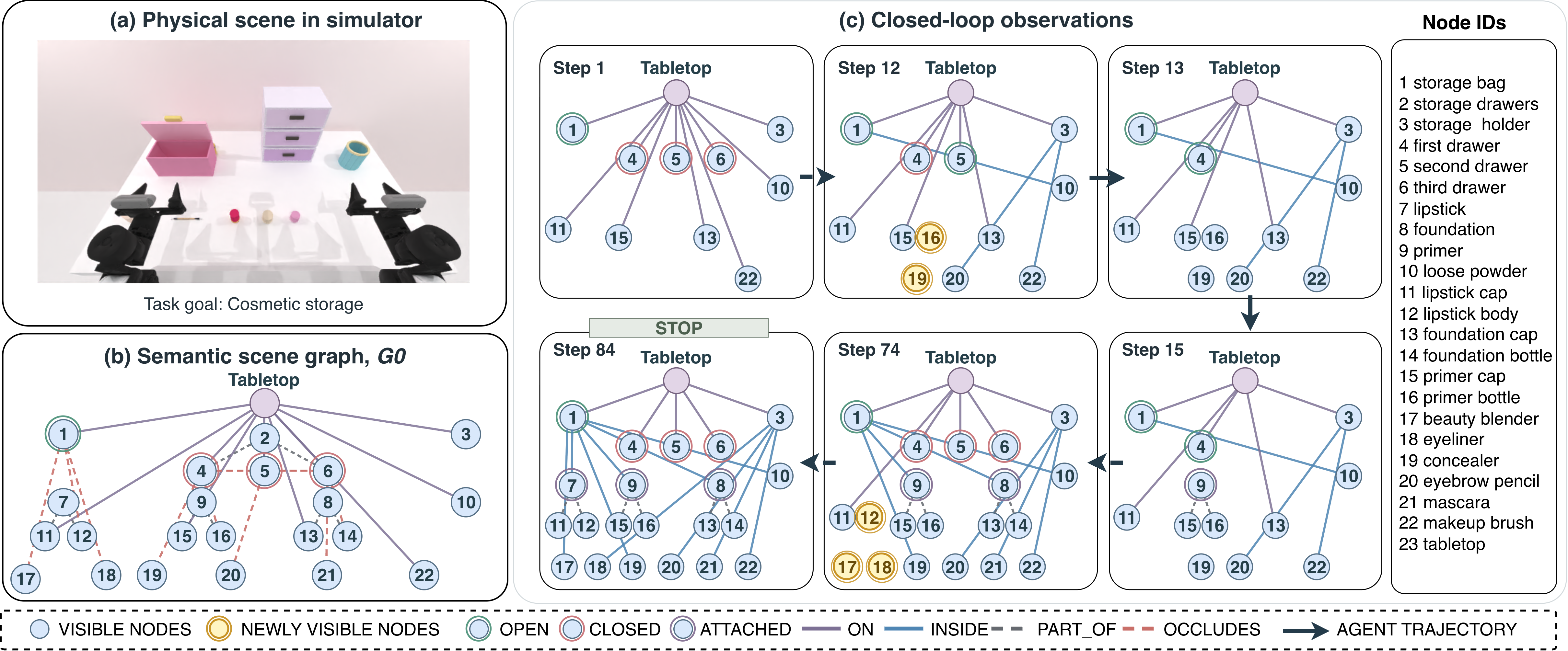} 
    \vspace{-1ex}
    \caption{
Visualization of instantiated \name during closed-loop evaluation. We illustrate a GPT-5.5 rollout on a cosmetic-storage task, in which the agent correctly terminates at step 84 after placing all objects in their locations. At each step, the agent observes only the currently visible subgraph and executes a symbolic action. At step 12, \texttt{open(first drawer)} reveals the concealer (NO.19) and primer bottle (NO.16); however, the opened first drawer occludes the already-open second drawer (NO.5), making it invisible in the next observation of step 13. Some objects like lipstick are composed of assemblable parts.
}
    \label{fig:vis}
    \vspace{-2ex}
\end{figure*}

\section{RoboGraph: A Task-State Horizon Compiler}

\paragraphb{Overview.}
As shown in \figurename~\ref{fig:overview}, \name is a task compiler that transforms a task-family specification and a source semantic scene graph into an executable task instance for closed-loop interaction. Compilation produces:
\begin{equation}
\mathcal{I}
=
\left(
G_0,g,\mathcal{A},O,\Lambda
\right),
\end{equation}
where $O$ specifies how the latent scene graph is compiled into the
agent's view, and $\Lambda$ specifies the exogenous events that may occur
during execution.

At run time, the compiled instance maintains the latent scene graph $G_t$,
compiles $(G_t,h_t)$ into the view graph $\widetilde{G}_t$ according to
Eq.~\eqref{eq:view}, and applies each agent action together with any
triggered event to update the world state according to
Eq.~\eqref{eq:trans}. The same instance can be executed by the symbolic
backend directly or mapped to simulator backends that implement the same
task-level action and event semantics.

\paragraph{Executable scene graph.}
\label{sec:scene-graphs}
\name{} represents each task instance as an executable semantic scene
graph.
Nodes represent scene entities, such as rooms, containers, surfaces, and objects. 
Each node stores its affordances and mutable state, which determine the applicable actions in $\mathcal{A}$ and record properties such as openness, location, and ownership. 
Edges represent spatial and functional relations, including containment, support, adjacency, and occlusion (see \figurename~\ref{fig:vis}).

The scene graph defines which facts are currently observable. 
Containment and occlusion may hide entities until enclosing containers are opened or blocking objects are moved aside, with nested constraints requiring sequential resolution. 
In indoor scenes, room boundaries further restrict visibility to entities in the agent’s current room.

Given a task-family specification, \name{} constructs the initial scene
graph $G_0$ and goal predicate $g$. 
The resulting scene structure and goal dependencies determine when task-relevant facts become observable, how long they remain relevant to subsequent decisions, and whether their values may change during execution. 
The TSH of each instance is therefore fixed during compilation.

\paragraph{Compiling partial observations.}
\label{sec:view-graph}

At each step, the observation operator derives the current view graph
from the world graph. 
Let
\begin{equation}
E_t^{\mathrm{block}}
=
\left\{
e\in E_t :
e \text{ blocks visibility under } X_t
\right\}
\end{equation}
denotes the active visibility constraints. 
The visible node set
$V_t^{\mathrm{vis}}$ includes entities in the agent's current room that are
reachable without crossing $E_t^{\mathrm{block}}$.

The observation is the induced view graph
\begin{equation}
O(G_t,h_t)
=
\left(
V_t^{\mathrm{vis}},
E_t[V_t^{\mathrm{vis}}],
\rho\!\left(X_t[V_t^{\mathrm{vis}}]\right)
\cup
\phi(h_t)
\right),
\label{eq:view-compile}
\end{equation}
where $E_t[V_t^{\mathrm{vis}}]$ retains relations among visible nodes,
$\rho$ selects observable attributes, and $\phi(h_t)$ provides feedback
from the preceding action and information revealed by inspection.

Actions affect subsequent observations by modifying $G_t$. 
Opening a
container, moving an occluder, or entering another room changes the next
view projection. 
Facts that leave the current view are removed rather than carried forward.
Visible changes therefore appear in the next view graph, while hidden changes remain unobserved until the affected region is revealed again. 
Failed actions leave $G_t$ unchanged and provide only failure feedback through $\phi(h_t)$.

\paragraph{Compiling state transitions.}
\label{sec:event-execution}

\name{} compiles the rules by which agent actions and exogenous events update the world graph from $G_t$ to $G_{t+1}$.
For each selected action $a_t$, \name{} checks its preconditions $\mathrm{Pre}(a_t)$ against the complete graph $G_t$ and applies its
effects $\mathrm{Eff}(a_t)$ as a graph rewrite. 
For example, \texttt{open(drawer)} requires the drawer to be closed and accessible, and changes its state from \textsc{closed} to \textsc{open}. If the preconditions are not satisfied, $G_t$ remains unchanged, and the failure outcome is returned. 
Exogenous changes are specified by an event schedule:
\begin{equation}
\Lambda=\{\lambda_1,\ldots,\lambda_K\},
\qquad
\lambda=
\langle
\psi_{\lambda},
\delta_{\lambda},
\kappa_{\lambda}
\rangle,
\end{equation}
where $\psi_{\lambda}$ is a trigger over the current execution state, $\delta_{\lambda}$ is the corresponding graph rewrite, and $\kappa_{\lambda}$ specifies when the rewrite is applied. 
Triggers are defined by task progress, such as a successful action or a newly satisfied goal fact, rather than by fixed time steps.
These events not only affect active task state, but also extend task goals.

\section{RoboGraphBench}

\paragraphb{Data analysis.}
Our benchmark contains 84 source scenes from seven task families, including 57 tabletop and 27 indoor scenes.
Each scene yields one baseline and six intervention conditions: action failure ($I_1$), state regression ($I_2$), completed-subgoal rollback ($I_3$), wrong relocation ($I_4$), added occlusion ($I_5$), and an added-object subgoal ($I_6$), producing 588 distinct episodes (399 tabletop and 189 indoor).
Across the scene graphs, each contains 18.4 nodes, 21.8 edges, 6.1 occluded nodes, and 14.3 subgoals on average.
Our final evaluation contains 8820 completed rollouts: 5985 tabletop rollouts and {2835} indoor rollouts across 15 models. More details of the analysis are provided in the {Appendix}.

\begin{table*}[t]
\centering
\resizebox{0.89\textwidth}{!}{%
\begin{tabular}{l *{8}{c} | *{8}{c}}
\toprule
\multirow{2}{*}{\textbf{Model}}
& \multicolumn{8}{c|}{\textbf{Tabletop (399 episodes, 57 scenes)}}
& \multicolumn{8}{c}{\textbf{Indoor (189 episodes, 27 scenes)}} \\
\cmidrule(lr){2-2}
\cmidrule(lr){3-4}
\cmidrule(lr){5-8}
\cmidrule(lr){9-9}
\cmidrule(lr){10-10}
\cmidrule(lr){11-12}
\cmidrule(lr){13-16}
\cmidrule(lr){17-17}
& $\overline{N}_s\downarrow$
& SR$\uparrow$
& SSAL$\uparrow$
& AER$\uparrow$
& RAR$\downarrow$
& SEG$\uparrow$
& ADR$\uparrow$
& IRR$\uparrow$
& $\overline{N}_s\downarrow$
& SR$\uparrow$
& SSAL$\uparrow$
& AER$\uparrow$
& RAR$\downarrow$
& SEG$\uparrow$
& ADR$\uparrow$
& IRR$\uparrow$ \\
\midrule

\addlinespace[2pt]
\textit{Closed-source models}
& \multicolumn{8}{c|}{}
& \multicolumn{8}{c}{} \\[2pt]

DeepSeek-V4 Pro
& 50.2 & 49.1 & 0.406 & 94.0 & 19.0 & 68.6 & 13.7 & 49.8
& 50.9 & 23.3 & 0.188 & 97.0 & 12.1 & 35.7 & 6.0 & 23.7 \\

GLM-5.2
& 55.3 & 43.6 & 0.324 & \underline{95.4} & 32.0 & 53.9 & 10.6 & 46.7
& 57.8 & 57.1 & 0.305 & 97.1 & 17.3 & \underline{43.7} & \underline{6.9} & \underline{57.6} \\

Qwen3.6-Plus
& 40.7 & 21.1 & 0.224 & 92.7 & 23.1 & 56.2 & 11.0 & 20.6
& 61.7 & 28.6 & 0.132 & \underline{98.7} & 14.5 & 32.7 & 5.7 & 30.9 \\

Qwen3.7-Plus
& 47.1 & 24.3 & 0.232 & 90.3 & 25.9 & 55.5 & 11.1 & 24.7
& 51.5 & 21.7 & 0.076 & 97.5 & 13.2 & 30.0 & 6.3 & 24.8 \\

GPT-5.4
& 69.5 & 23.1 & 0.190 & 93.6 & 47.7 & 40.7 & 8.3 & 25.1
& 52.0 & 41.8 & 0.283 & 95.4 & 16.7 & 35.1 & 6.2 & 44.2 \\

Claude Opus 4.7
& 48.8 & 56.6 & 0.439 & 89.2 & 24.4 & \underline{69.1} & \underline{14.8} & 57.4
& 44.5 & \underline{80.4} & \underline{0.553} & 96.9 & 13.2 & 40.7 & \textbf{8.6} & \textbf{81.4} \\

Gemini 3.1 Pro
& \underline{40.2} & \underline{62.7} & \underline{0.529} & 94.1 & \underline{13.3} & 67.6 & 14.7 & \underline{63.4}
& \textbf{35.9} & 42.3 & 0.371 & 98.3 & \underline{10.9} & 32.7 & 5.5 & 42.8 \\

GPT-5.5
& \textbf{40.0} & \textbf{78.4} & \textbf{0.689} & \textbf{97.4} & \textbf{12.2} & \textbf{76.9} & \textbf{16.7} & \textbf{78.6}
& \underline{41.5} & \textbf{82.5} & \textbf{0.607} & \textbf{99.4} & \textbf{10.6} & \textbf{44.5} & \textbf{8.6} & \textbf{81.4} \\

\midrule
\addlinespace[2pt]
\textit{Open-source models}
& \multicolumn{8}{c|}{}
& \multicolumn{8}{c}{} \\[2pt]

Qwen3-VL-8B
& 42.1 & 1.5 & 0.003 & 69.7 & 46.7 & 28.7 & 4.7 & 2.4
& 50.9 & 0.5 & 0.000 & 59.1 & 41.0 & 13.1 & 2.3 & 1.1 \\

Qwen3-VL-32B
& 71.7 & 9.3 & 0.074 & 81.6 & 51.6 & 45.0 & 8.3 & 11.6
& 61.1 & 9.0 & 0.034 & 88.4 & 24.8 & 35.5 & \underline{6.9} & 11.3 \\

InternVL3.5-8B
& 96.0 & 2.3 & 0.003 & 31.3 & 76.9 & 29.4 & 4.8 & 3.5
& 98.5 & 0.5 & 0.000 & 26.1 & 77.6 & 12.7 & 3.5 & 1.2 \\

InternVL3.5-38B
& 90.7 & 6.3 & 0.033 & 32.3 & 74.1 & 30.0 & 5.1 & 8.4
& 84.9 & 5.8 & 0.039 & 63.0 & 44.7 & 32.2 & 6.1 & 6.5 \\

Cosmos-Reason1-7B
& 79.0 & 0.0 & 0.000 & 22.6 & 63.6 & 17.2 & 3.1 & 0.0
& 93.1 & 0.0 & 0.000 & 12.8 & 71.7 & 6.6 & 1.3 & 0.0 \\

RoboBrain2.0-7B
& 58.3 & 0.0 & 0.000 & 38.1 & 68.7 & 22.0 & 3.8 & 0.0
& 80.5 & 0.0 & 0.000 & 11.5 & 78.5 & 3.9 & 0.2 & 0.0 \\

RoboBrain2.0-32B
& 64.6 & 0.3 & 0.000 & 51.6 & 63.7 & 24.7 & 4.2 & 0.5
& 54.5 & 0.0 & 0.000 & 48.8 & 54.7 & 13.6 & 3.1 & 0.0 \\

\bottomrule
\end{tabular}%
}
\vspace{-1ex}
\caption{
Performance of 15 agentic models under semantic closed-loop evaluation
across tabletop and indoor task families. Every model is evaluated on
399 tabletop episodes and 189 navigation-enabled indoor episodes.
Best and second-best displayed values in each column are bolded and
underlined, respectively; ties receive the same formatting.
}
\label{tab:models}
\vspace{-3ex}
\end{table*}

\begin{figure*}[t]
\centering
\includegraphics[width=0.9\textwidth]{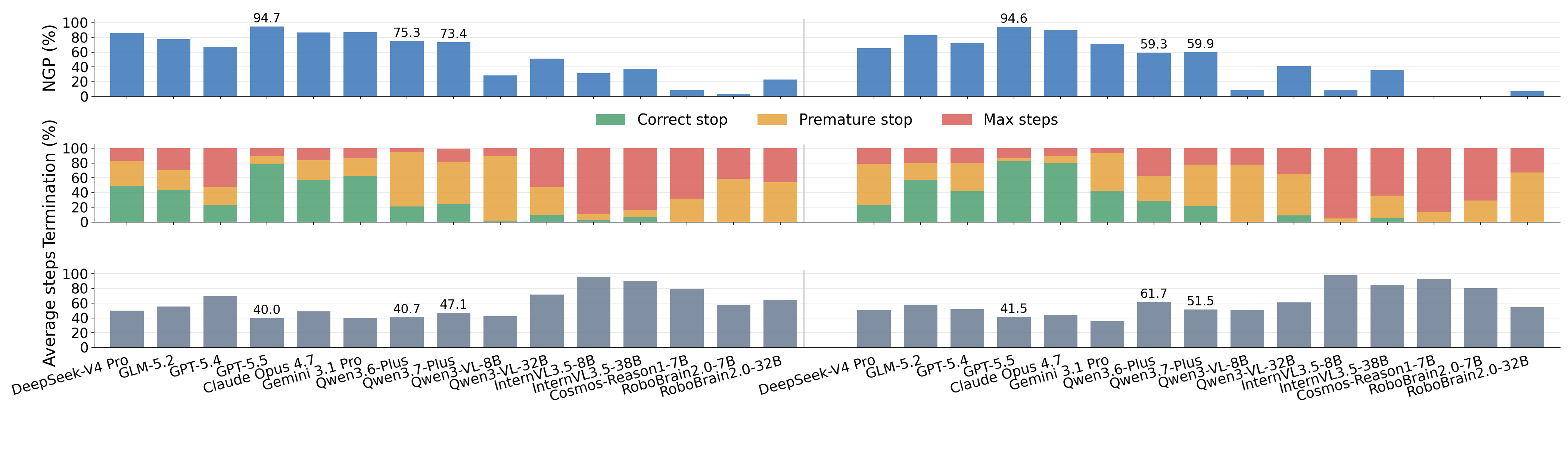}
\vspace{-6ex}
\caption{
For each model, we report (1) Normalized Goal Progress (NGP), the normalized fraction of task subgoals completed; (2) termination distribution, the proportions of episodes that terminate correctly, prematurely, or neither; and (3) average steps, the average number of agent execution steps per trajectory. Results are reported separately for tabletop and indoor tasks.
}
\vspace{-2ex}
\label{fig:outcome-diagnostics}
\end{figure*}

\section{Experiments}

\subsection{Setup}
\paragraphb{Base models.}
We evaluate 15 foundation models, including 8 closed-source models comprising 2 LLMs, DeepSeek-V4 Pro~\cite{xu2026deepseek} and GLM-5.2~\cite{zeng2026glm}, and 6 VLMs, GPT-5.4~\cite{openai2026gpt54}, GPT-5.5~\cite{openai2026gpt55}, Claude Opus 4.7~\cite{anthropic2026claudeopus47}, Gemini 3.1 Pro~\cite{deepmind2026gemini31pro}, Qwen3.6-Plus~\cite{qwen36plus}, and Qwen3.7-Plus~\cite{qwen37plus}.
Also, we evaluate 7 open-source models, covering 4 general-purpose VLMs, Qwen3-VL-8B/32B~\cite{bai2025qwen3}, InternVL3.5-8B/38B~\cite{wang2025internvl3}, and 3 VLMs for robotics, Cosmos-Reason1-7B~\cite{azzolini2025cosmos} and RoboBrain2.0-7B/32B~\cite{team2025robobrain}.
These models span diverse families, scales, and modalities, enabling a comprehensive evaluation across model capabilities.
Note that in our semantic evaluation setting, visual observations are replaced with symbolic graphs, allowing text-only LLMs to operate as closed-loop embodied agents.

\paragraphb{Closed-loop evaluation.}
We deploy all models within a unified agentic harness for closed-loop evaluation.
At each step, the agent receives the current observation, final task goal, available action skills, and a history of its 16 most recent actions as context.
The agent then selects an action, and repeatedly observes and acts until it emits \texttt{stop} or reaches the maximum step limit, e.g., 100 steps in our experiments, which largely exceeds the length of oracle trajectories.
When an action fails, the harness returns the corresponding feedback, enabling failure recovery.
Additional details of prompt and context are provided in the {Appendix}.

Our harness supports two observation interfaces: symbolic scene graph and robot-camera images, together with their corresponding execution backends.
In the \textit{symbolic backend}, the agent observes only the currently visible view graph rather than the complete latent scene graph, preserving partial observability analogous to real-world perception.
As actions are executed, newly observable nodes and relations are incrementally added to the visible graph.
In the \textit{simulation backend}, the agent receives multiple-view camera images rendered by the simulator, i.e., RoboTwin 2.0~\cite{chen2025robotwin} for tabletop tasks and RoboCasa~\cite{nasiriany2024robocasa} for indoor kitchen tasks.

\paragraphb{Metrics.}
To assess agent state-management performance across varying task-state horizons (TSHs), we introduce a suite of complementary metrics.
First, to evaluate overall task completion and execution quality, we use 2 metrics:
\begin{itemize}
\item \textbf{Task Success Rate (SR, \%):} the proportion of episodes that achieve the task goal and terminate correctly.
\item \textbf{Success weighted by Symbolic Action Length (SSAL):} the success-weighted ratio of the symbolic oracle's shortest action-sequence length ($L_o$) to the agent's number of non-stop action steps ($N_s$). 
Detailed formulations are provided in the Appendix.
\end{itemize}
Second, to assess state-management capability, we introduce 5 metrics:
\begin{itemize}
\item \textbf{Action Executability Rate (AER, \%; maintenance):} the proportion of executable actions whose preconditions hold in the current world state. 
\item \textbf{Redundant Action Rate (RAR, \%; maintenance):} the percentage of unnecessary actions that repeat a previously attempted action.
\item \textbf{State Exploration Gain (SEG, \%; exploration):} the percentage of task-relevant spatial dependencies revealed by successful exploratory actions.
\item \textbf{Action-Distance Reduction Rate (ADR, \%; exploration):} the average reduction in the symbolic oracle's remaining ordered action distance, measuring how effectively exploration resolves prerequisite dependencies.
\item \textbf{Intervention Recovery Rate (IRR, \%; updating):} the proportion of intervention-induced state updates that the agent detects and successfully recovers from.
\end{itemize}
Besides, we leverage two metrics to measure the ability of failure recovery:
\begin{itemize}
\item \textbf{F1@Recovery Detection (F1@RD, \%):} the F1 score for detecting when failure recovery is required.
\item \textbf{Acc@Recovery Grounding (Acc@RG, \%):} the accuracy of grounding the failed action and relevant object.
\end{itemize}

\subsection{Research Questions}
\paragraphb{RQ1:} How does agent performance vary across the three dimensions of state management: maintenance, exploration, and updating?

\paragraphb{RQ2:} How does the proposed TSH relate to conventional action-sequence length and subgoal complexity in characterizing long-horizon difficulty?

\paragraphb{RQ3:} How does recovery difficulty vary across dynamic interventions, and which severely disrupt task completion?


\subsection{Main Results}

\paragraphb{Overall performance varies across models (RQ1).}
As shown in Table~\ref{tab:models}, GPT-5.5 achieves the strongest overall performance across both task families, reaching an SR of {78.4\%} on tabletop tasks and {82.5\%} on indoor tasks. 
It also attains the highest SSAL, indicating that it completes tasks successfully with action sequences that are comparatively close to the symbolic oracle.
Gemini 3.1 Pro ranks second in tabletop SR at {62.7\%}, while Claude Opus 4.7 approaches GPT-5.5 on indoor tasks with an SR of {80.4\%}.
By contrast, Qwen3.6-Plus and Qwen3.7-Plus perform substantially worse on these tasks, achieving SRs of no more than {30\%}, revealing considerable limitations on tasks with demanding TSHs. 
The performance gap between closed- and open-source models is even more pronounced. 
Among the open-source models, Qwen3-VL-32B performs best on tabletop tasks, yet achieves an SR of only 9.3\%, while Cosmos-Reason1-7B and RoboBrain2.0-7B fail to complete any task.

To better understand these performance gaps, we analyze Normalized Goal Progress (NGP), termination distributions, and average step counts. 
As shown in \figurename~\ref{fig:outcome-diagnostics}, \textit{different models fail in different ways}. 
Qwen3.6-Plus and Qwen3.7-Plus achieve NGPs of approximately 75\% on tabletop tasks and 60\% on indoor tasks, yet frequently terminate before completing the goal, \textit{indicating insufficient exploration of unresolved task-relevant states}.
Gemini 3.1 Pro uses only 35.9 average steps on indoor tasks, fewer than GPT-5.5's 41.5, but its SR is 40.2 points lower.
By contrast, GPT-5.4 reaches the maximum step limit in 52.6\% of tabletop rollouts, and open-source models frequently exhaust the step budget, e.g., InternVL3.5-8B and InternVL3.5-38B average 96.0 and 90.7 steps per rollout, \textit{suggesting prolonged but unproductive exploration and poor maintenance}.

\begin{figure}[t]
    \centering
    \includegraphics[width=0.9\linewidth]{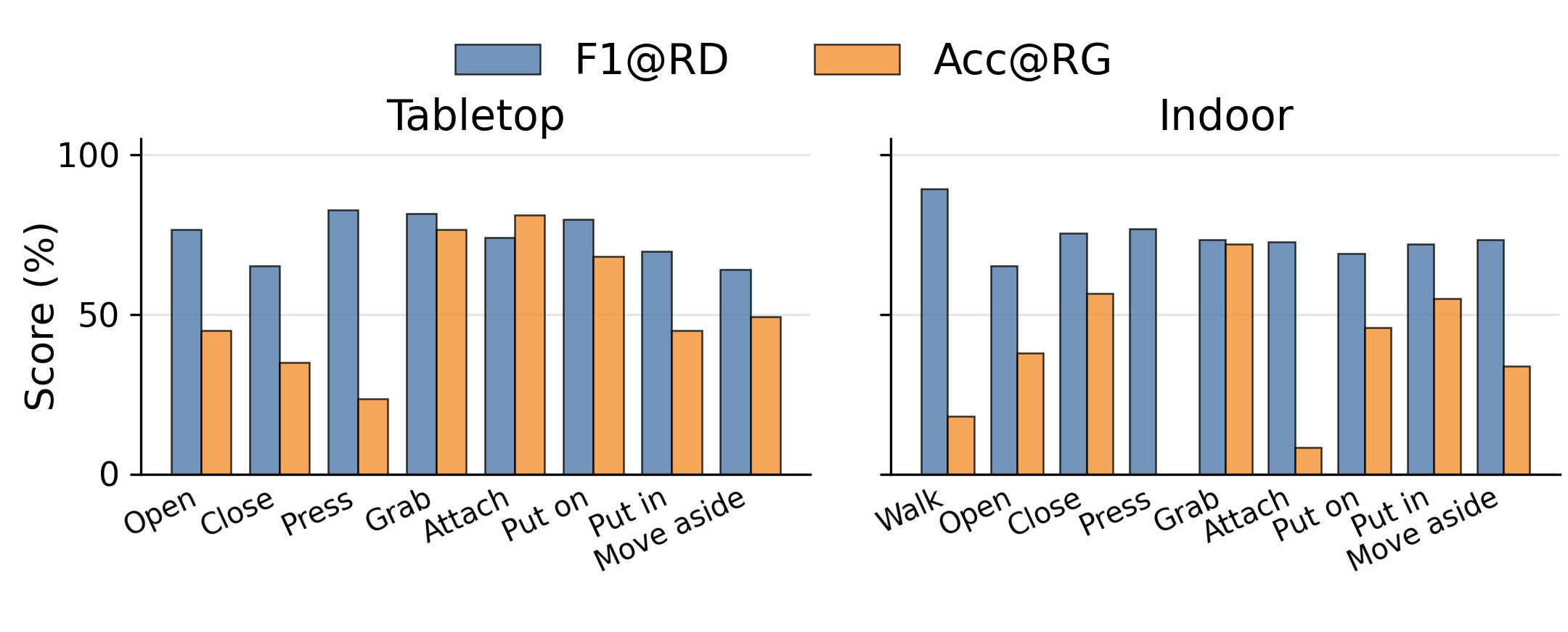}
    \vspace{-3ex}
    \caption{Recovery detection and grounding performance by action type on tabletop and indoor tasks, measured by F1@RD and Acc@RG, respectively.}
    \label{fig:recovery-by-action}
    \vspace{-2ex}
\end{figure}

\begin{figure}[t]
\centering
\includegraphics[width=0.85\linewidth]{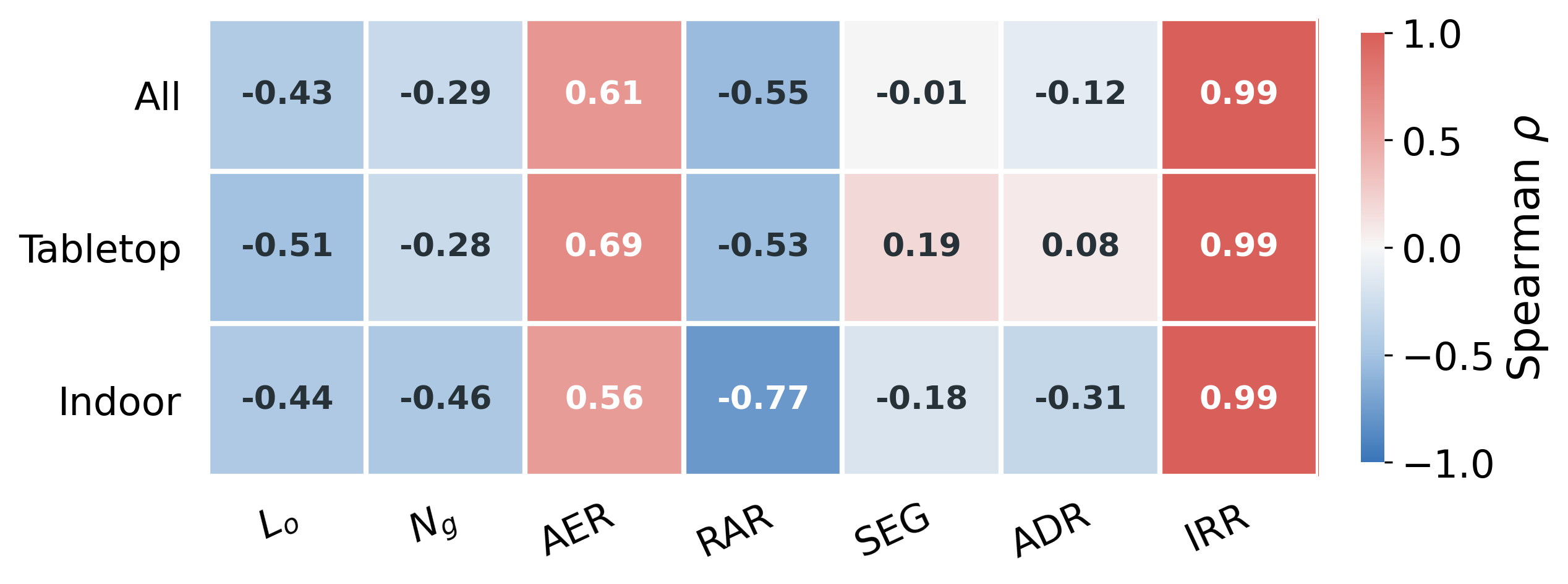}
\vspace{-2ex}
\caption{Between-scene correlations of oracle action-sequence length ($L_o$), subgoal counts ($N_g$), and 5 proposed state-management metrics with task success rate.}
\label{fig:scene-correlation}
\vspace{-2ex}
\end{figure}

\begin{figure}[t]
\centering
\includegraphics[width=0.7\linewidth]{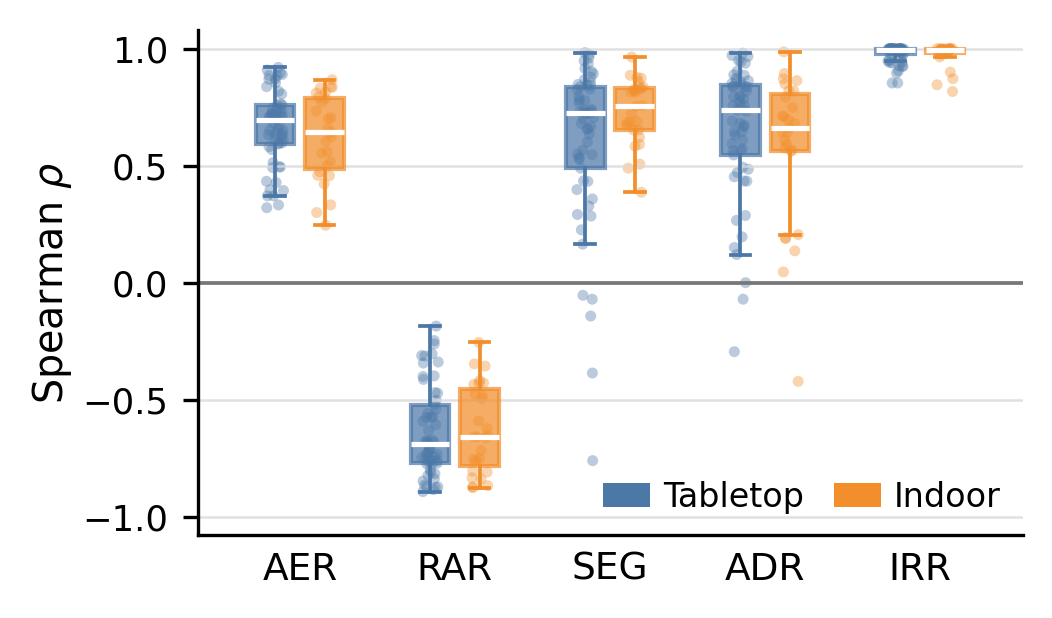}
\vspace{-2ex}
\caption{Within-scene correlations of 5 proposed state-management metrics with task success rate across the models evaluated on that scene.}
\label{fig:within-scene-correlation}
\vspace{-2ex}
\end{figure}

\begin{table}[t]
\centering
\small
\setlength{\tabcolsep}{4pt}
\resizebox{0.65\linewidth}{!}{%
\begin{tabular}{@{}clrrr@{}}
\toprule
\textbf{Intervention}
& \textbf{Domain}
& $\Delta$SR (pp)
& IRR (\%)
& Latency \\
\midrule

\multirow{2}{*}{$I_1$}
& Tabletop & $-1.9$ & 26.4 & {35.8} \\
& Indoor   & $-0.7$ & 28.0 & {37.1} \\
\addlinespace[2pt]

\multirow{2}{*}{$I_2$}
& Tabletop & $+0.1$ & 33.7 & 22.6 \\
& Indoor   & $-1.0$ & 31.0 & 28.1 \\
\addlinespace[2pt]

\multirow{2}{*}{$I_3$}
& Tabletop & $-2.7$ & 28.9 & 30.3 \\
& Indoor   & $-1.5$ & 30.6 & 29.5 \\
\addlinespace[2pt]

\multirow{2}{*}{$I_4$}
& Tabletop & $+0.9$ & 31.5 & 32.4 \\
& Indoor   & $-1.0$ & 32.0 & 30.5 \\
\addlinespace[2pt]

\multirow{2}{*}{$I_5$}
& Tabletop & $-10.4$ & 27.6 & 27.5 \\
& Indoor   & $-4.4$  & 34.3 & 23.9 \\
\addlinespace[2pt]

\multirow{2}{*}{$I_6$}
& Tabletop & $-4.3$ & 41.0 & 19.4 \\
& Indoor   & $-1.2$ & 43.3 & 24.0 \\

\bottomrule
\end{tabular}}
\vspace{-1ex}
\caption{
Recovery performance by intervention type. $\Delta$SR compares each
intervention with its paired baseline, and latency is averaged
over successfully recovered rollouts.
}
\vspace{-3ex}
\label{tab:rq3-intervention}
\end{table}

\paragraphb{State-management capabilities vary across models (RQ1).}
For \textit{state maintenance}, GPT-5.5 achieves the highest AER and the lowest RAR across both task families, with an AER/RAR of 97.4\%/12.2\% on tabletop tasks and 99.4\%/10.6\% on indoor tasks. 
Across models, \textit{achieving high action executability appears relatively easy, whereas avoiding redundant actions remains challenging}.
This discrepancy suggests that agents can often select actions whose preconditions are satisfied, yet struggle to leverage history to recall prior action outcomes and maintain an accurate representation of the current task state. 
GPT-5.4 exemplifies this on tabletop tasks: despite achieving a 93.6\% AER, it repeatedly revisits previously attempted actions, resulting in a 47.7\% RAR and an SR of only 23.1\%.

For \textit{state exploration}, GPT-5.5 performs best, achieving 76.9\% SEG and 16.7\% ADR on tabletop tasks, while Claude Opus 4.7 and Gemini 3.1 Pro also achieve competitive results.
Nearly all models exhibit {lower SEG and ADR} on indoor tasks than on tabletop tasks, whereas AER and RAR remain broadly comparable across the two domains. 
This gap reflects the greater resolving demands of spatial and temporal dependencies for indoor exploration, where agents must first identify the correct room and then reveal the target object. 
Thus, \textit{selecting locally executable actions alone is insufficient to assess state management}.

For \textit{state updating and recovery}, GPT-5.5 reaches 78.6\% IRR for tabletop, while GPT-5.5 and Claude Opus 4.7 both reach 81.4\% for indoor.
One intervention type injects execution-time failures into randomly selected actions.
We report F1@RD and Acc@RG disaggregated by action type in \figurename~\ref{fig:recovery-by-action}. 
Recovery detection is generally easier than issuing accurate recovery action.

\paragraphb{State-management metrics disentangle TSH difficulty from model capability (RQ2).}
Figure~\ref{fig:scene-correlation} presents a \textit{between-scene} analysis. We average results across models within each scene and then compare these scene-level averages across scenes, indicating which metrics are associated with a scene being more difficult for models overall.
Oracle action-sequence length and subgoal count are only moderately correlated with SR. 
In comparison, AER is most informative on tabletop tasks ($\rho=0.69$), whereas RAR is most informative on indoor tasks ($\rho=-0.71$). 
SEG and ADR exhibit relatively weak scene-level correlations, suggesting that they are less indicative of distinguishing the relative difficulty of different scenes.
Figure~\ref{fig:within-scene-correlation} instead presents a \textit{within-scene} analysis. 
We fix one scene and compute correlations across models, revealing which metrics distinguish successful from unsuccessful models.
In this setting, SEG and ADR are strongly associated with SR on both tabletop tasks, with median correlations of $0.72$ and $0.74$, and indoor tasks, with median correlations of $0.61$ and $0.56$. Thus, exploration metrics clearly distinguish which models handle the same scene more effectively. 
In summary, action length and subgoal count characterize coarse task difficulty, whereas state-management metrics reveal complementary differences in model capability in varying TSHs.

\paragraphb{Occlusion is the most disruptive intervention (RQ3).}
As shown in Table~\ref{tab:rq3-intervention}, added occlusion causes the largest SR decline in both task families, reducing performance by 10.4 percentage points on tabletop tasks and 4.4 points indoors. 
This suggests that interventions that introduce new spatial and temporal dependencies impose particularly demanding state-updating requirements. 
Action failures incur the longest recovery latency, at approximately 36 steps, indicating that recovery requires not only detecting the state change but also repairing its consequences. Note that IRR is computed only over rollouts in which an intervention is actually triggered, whereas $\Delta$SR compares each intervention against its paired no-intervention baseline.


\begin{figure}[t]
\centering
\includegraphics[width=0.95\linewidth]{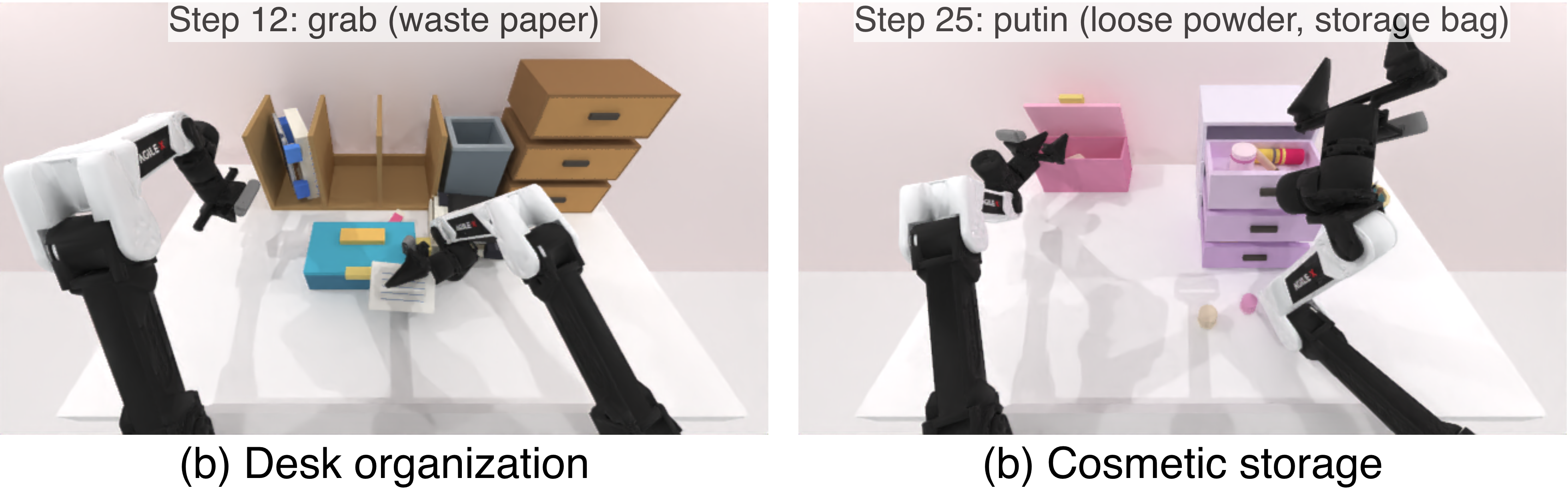}
\vspace{-1ex}
\caption{Visual observation during closed-loop evaluation.}
\label{fig:robotwin}
\vspace{-2ex}
\end{figure}

\subsection{Extension to Visual Backends}
\name also supports simulator backends.
We evaluate 8 closed-source models on two matched RoboTwin tabletop tasks (see \figurename~\ref{fig:robotwin}): \textit{desk organization} and \textit{cosmetics storage}, using three-view RGB observations from the head, right, and left camera, instead of symbolic view graphs.
All models are evaluated using the same unified harness as in the semantic setting.
We find that all metrics in the semantic and visual settings are positively correlated, e.g., Spearman's $\rho=0.37$ for NGP and $\rho=0.75$ for ADR.
On the more challenging \textit{cosmetics storage} task, where performance varies more widely across models, the correlation increases to $\rho=0.63$.
Task success rates are substantially lower in simulation, likely reflecting the additional demands of visual grounding and state decoding. 
These results suggest that symbolic view graphs primarily isolate planning and state-management quality, whereas simulation additionally evaluates perception and precise physical execution.


\section{Conclusion}
In this paper, we introduce a complementary perspective on long-horizon difficulty in embodied tasks: the span over which an agent must maintain known state, explore hidden state, and update stale state. 
We formalize this span as the \textit{task-state horizon} (TSH), develop \name{} to compile controlled TSHs into executable tasks, and construct \benchname{} for closed-loop evaluation. 
Our systematic analysis of {8820} rollouts across tabletop and indoor domains and 15 foundation models shows that TSH-based state-management measures are more closely associated with task success than conventional proxies based on action-sequence length or subgoal count. 
We hope that TSH-controlled task compilation will become a standard framework for evaluating and training embodied agents that must act reliably in a dynamic and evolving world.

\bibliography{aaai2027}

\clearpage
\appendix

\section{Benchmark and Data Analysis}
\label{app:benchmark-analysis}

\paragraphb{Compiled task instances.}
As mentioned in the paper, \name compiles a task-family specification and a source scene graph into an
executable instance
\begin{equation}
    \mathcal{I}=(G_0,g,\mathcal{A},\Lambda),
    \label{eq:compiled-instance}
\end{equation}
where $G_0$ is the initial latent scene graph,$g$ is goal predicates, $\mathcal{A}$ is the action skill, and $\Lambda$ is an event schedule.  

\paragraphb{Dataset composition.}
Table~\ref{tab:appendix-data-composition} reports the source-scene and episode
coverage.  
The 84 source scenes instantiate one baseline and six intervention
conditions each, yielding 588 independent episodes. 
Across all scene graphs, there are 1544 node entities and 1829 directed edge relations.  
On average, a graph contains $18.4\pm3.3$ nodes, $21.8\pm6.1$ edges, and $6.1\pm1.8$ initially occluded nodes.  
The task specifications contain 1200 logical goal predicates, corresponding to $14.3\pm4.1$ subgoal counts per scene.
The average symbolic oracle action-sequence length is $21.7\pm5.4$ actions across the 84 source scenes.
  
Given an initial state, we compute the ordered action distance used in ADR as
\[
C_{\mathrm{distance}}(G_0)
=
C_{\mathrm{goal}}(G_0)
+
C_{\mathrm{pre}}(G_0),
\]
where $C_{\mathrm{distance}}(G_0)$ counts the actions that directly satisfy
\textit{currently unsatisfied goal predicates}, and $C_{\mathrm{pre}}(G_0)$ counts
\textit{the prerequisite actions required to access the relevant entities and
execute those goal-directed actions}. 
Such prerequisites include navigation, opening closed containers, removing occluders, and acquiring objects for manipulation. 
Prerequisites shared by multiple goal predicates are counted only once.
For example, consider a tabletop task with the goals
\[
\mathrm{INSIDE}(\text{red pen},\text{pencil case})
\land
\mathrm{CLOSED}(\text{pencil case}).
\]
Suppose that the red pen is initially occluded by a book, the pencil case
is open, and both objects are located in the agent's current room.
A valid action sequence is
\[
\begin{aligned}
\pi = [&\operatorname{move\_aside}(\text{book}),
       \operatorname{grab}(\text{red pen}),\\
      &\operatorname{putin}(\text{red pen},\text{pencil case}),
       \operatorname{close}(\text{pencil case})].
\end{aligned}
\]
The initial action distance is therefore
\[
C_{\mathrm{distance}}(G_0)=1+1+1+1=4.
\]
If both the red pen and the pencil case are instead located in an adjacent room, one additional $\operatorname{walk}$ action is required, yielding $C_{\mathrm{distance}}(G_0)=5$. 
If the pencil case is initially closed, an additional $\operatorname{open}$ action must be executed before $\operatorname{putin}$; the case must subsequently be closed again to satisfy the $\mathrm{CLOSED}$ predicate.
This ordered interaction exemplifies a temporal dependency.
Across all 84 source scenes, the average ordered action distance is $24.3\pm5.6$. 

\begin{table}[t]
\centering
\resizebox{\columnwidth}{!}{%
\begin{tabular}{lrrrr}
\toprule
\textbf{Task family} & \textbf{Tabletop} & \textbf{Indoor}
& \textbf{Sources} & \textbf{Episodes} \\
\midrule
Buffet-tray sorting           & 7  & 7  & 14 & 98  \\
Cosmetics storage             & 10 & 0  & 10 & 70  \\
Desk organization             & 10 & 0  & 10 & 70  \\
Dining-table organization     & 10 & 0  & 10 & 70  \\
Food packing   & 0  & 10 & 10 & 70  \\
Fridge-condiment loading      & 10 & 10 & 20 & 140 \\
Toy organization              & 10 & 0  & 10 & 70  \\
\midrule
\textbf{Total}                & \textbf{57} & \textbf{27}
& \textbf{84} & \textbf{588} \\
\bottomrule
\end{tabular}%
}
\caption{Task composition in \benchname. Each task comprises one base scene and six mutually exclusive intervention conditions derived from that scene.}
\label{tab:appendix-data-composition}
\vspace{-2ex}
\end{table}

\paragraphb{Event schedule.}
Dynamic changes are compiled into
\begin{equation}
    \Lambda=\{\lambda_1,\ldots,\lambda_K\},\qquad
    \lambda=\langle\psi_\lambda,\delta_\lambda,\kappa_\lambda\rangle ,
    \label{eq:event-schedule}
\end{equation}
where $\psi_{\lambda}$ is a trigger over the current execution state, $\delta_{\lambda}$ is the corresponding graph rewrite, and $\kappa_{\lambda}$ specifies when the rewrite is applied. 
Triggers refer to semantic progress, such as the first
normally executable action of a type or the first satisfied placement
fact, rather than by fixed time steps.

Table~\ref{tab:appendix-interventions} details the six event types.  
``Triggered rollouts'' counts evaluated rollouts in which the event fired at least once, out of 855 tabletop and 405 indoor intervention rollouts per event type.
This rollout-level convention is important for $I_1$: its trigger may fire once for each distinct action type within the same rollout, whereas $I_2$--$I_6$ are applied at most once per rollout.

\begin{table*}[t]
\centering
\scriptsize
\setlength{\tabcolsep}{3.2pt}
\begin{tabular}{@{}clp{0.205\textwidth}p{0.325\textwidth}rrr@{}}
\toprule
& & & & \multicolumn{3}{c}{\textbf{Triggered rollouts}} \\
\cmidrule(lr){5-7}
\textbf{ID} & \textbf{Event} & \textbf{Trigger $\psi_\lambda$}
& \textbf{State update $\delta_\lambda$}
& \textbf{T / 855} & \textbf{I / 405} & \textbf{All / 1260} \\
\midrule
$I_1$ & Action failure
& First otherwise successful execution of each action type; at most once per
action name.
& Suppress the action effect, leave $G_t$ unchanged, and return typed failure
feedback.  The agent must retry or replan.
& 840 & 389 & 1,229 \\
\addlinespace[2pt]
$I_2$ & State regression
& First achieved a reversible node state, such as an opened container, after the minimum-step guard.
& Reset one node attribute (e.g., \texttt{open}, \texttt{closed}, or
\texttt{moved\_aside}) to its prior value.
& 689 & 348 & 1,037 \\
\addlinespace[2pt]
$I_3$ & Completed-subgoal rollback
& Currently satisfied object-placement goal fact.
& Remove the satisfied \texttt{ON}/\texttt{INSIDE} relation and relocate the
object to a valid non-goal staging surface.
& 744 & 346 & 1,090 \\
\addlinespace[2pt]
$I_4$ & Wrong relocation
& Currently satisfied object-inside-the-container goal fact.
& Update the object's location relative to another visible, plausible, but wrong container.
& 731 & 338 & 1,069 \\
\addlinespace[2pt]
$I_5$ & Added occlusion
& Eligible mid-episode blocker--target pair whose occlusion is
resolvable by an available action schema.
& Move a goal-relevant object behind a blocker and add an active
\texttt{OCCLUDES} edge; recovery may require \texttt{open} or
\texttt{move\_aside} followed by placement restoration.
& 518 & 108 & 626 \\
\addlinespace[2pt]
$I_6$ & Added object
& A source object's local goal becomes satisfied, or a designated container
reaches its capacity condition.
& Add an object node and staging relation, then extend the active goal by an
inherited placement rule (74 scenes) or activate an existing capacity-bound
goal (10 scenes).
& 490 & 245 & 735 \\
\bottomrule
\end{tabular}
\caption{Intervention semantics and observed trigger coverage over the full
8820-rollout evaluation.  T and I denote tabletop and indoor, respectively.
Counts are the numbers of rollouts with at least one applied intervention.}
\label{tab:appendix-interventions}
\vspace{-2ex}
\end{table*}

\begin{figure*}[t]
\centering
\includegraphics[width=\textwidth]{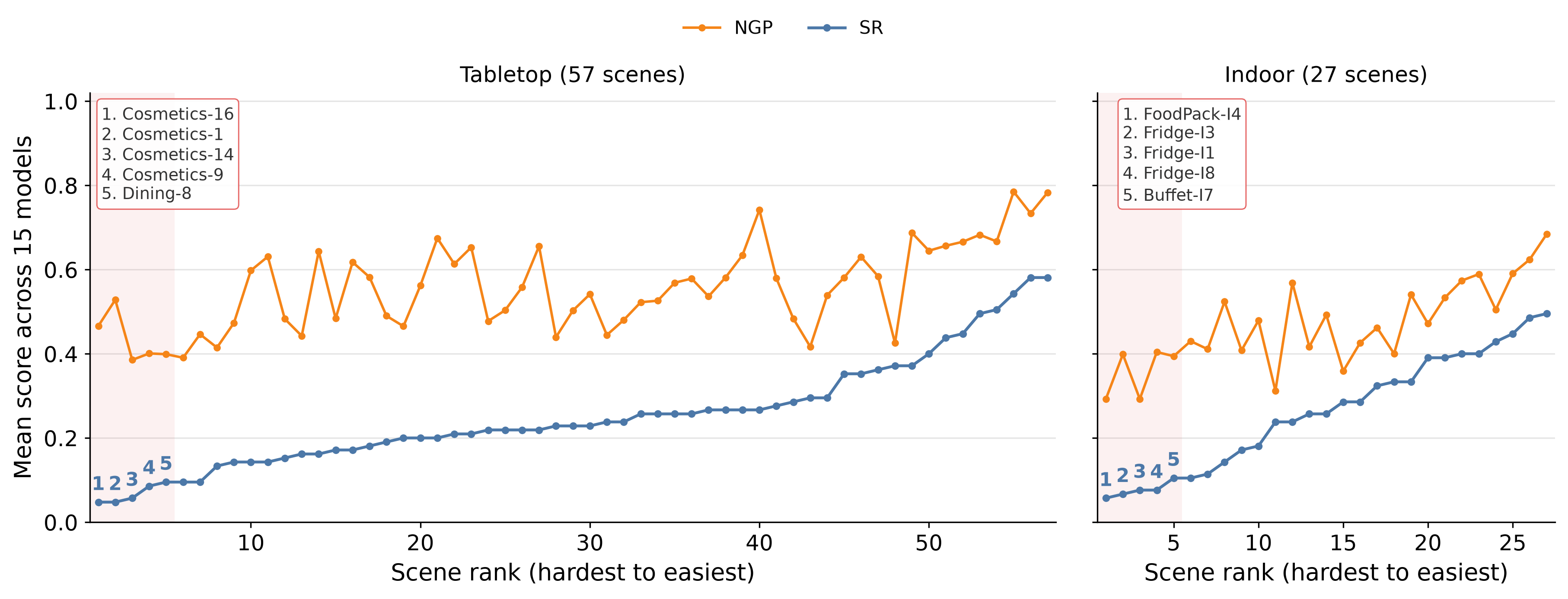}
\caption{Scenes ranked from hardest to easiest by mean SR over 15 models and
seven conditions.  NGP reveals that many low-SR scenes still permit partial
progress.  The shaded region and numbered key identify the five hardest
scenes in each domain.}
\label{fig:appendix-scene-difficulty}
\vspace{-2ex}
\end{figure*}

\begin{table*}[t]
\centering
\scriptsize
\setlength{\tabcolsep}{3.0pt}
\begin{tabular}{@{}lp{0.09\textwidth}p{0.36\textwidth}ccp{0.13\textwidth}c@{}}
\toprule
\textbf{Action} & \textbf{Parameters} & \textbf{Description}
& \textbf{Free} & \textbf{Held} & \textbf{Hand result}
& \textbf{FI} \\
\midrule
\texttt{observe} & --- &
Refresh the current observation without changing object state.
& 0 & No & hands unchanged & No \\
\texttt{walk}$^\dagger$ & target &
Move to an adjacent room in a navigation-enabled indoor scene.
& 0 & No & hands unchanged & Yes \\
\texttt{open} & container &
Open a visible closed openable container or reveal-capable object.
& 2 & No & hands unchanged & Yes \\
\texttt{close} & container &
Close a visible open openable container.
& 2 & No & hands unchanged & Yes \\
\texttt{press} & object &
Press a visible reachable object with \texttt{PRESSABLE} affordance.
& 1 & No & hands unchanged & Yes \\
\texttt{grab} & object &
Try to hold a visible reachable grabbable movable object; loaded carriers
may reject the attempt.
& 1 & No & occupies one hand & Yes \\
\texttt{attach} & part, target &
Attach two visible or held compatible part nodes; use this action to satisfy
\texttt{ASSEMBLED} goals.
& 2 & No & two-hand coordination & Yes \\
\texttt{puton} & object, target &
Place a held object on a visible surface without creating a self-placement
or placement cycle.
& 0 & Yes & frees one hand & Yes \\
\texttt{putin} & object, container &
Place a held object inside a visible container that is open if openable,
without creating a self-containment or containment cycle.
& 0 & Yes & frees one hand & Yes \\
\texttt{move\_aside} & object &
Move a visible movable blocker aside so hidden nodes may become visible.
& 1 & No & hands unchanged & Yes \\
\texttt{stop} & --- &
End the episode and submit the current state for success evaluation.
& 0 & No & hands unchanged & No \\
\bottomrule
\end{tabular}
\vspace{2pt}
\parbox{0.96\textwidth}{\scriptsize
\textbf{Free} is \texttt{required\_free\_hands}; \textbf{Held} is
\texttt{held\_object\_required}; \textbf{FI} is
\texttt{failure\_injectable}.  $^\dagger$\texttt{walk} is included only for
navigation-enabled indoor scenes.}
\caption{Complete action catalog exposed by the closed-loop evaluation harness.  
Each row corresponds to the serialized fields of one catalog entry: description, parameters, hand-usage, and injectable failure.}
\label{tab:appendix-action-catalog}
\vspace{-2ex}
\end{table*}

\section{Closed-Loop Evaluation Harness}
\label{app:closed-loop-harness}

\paragraphb{Interaction protocol.}
All models run in the same agentic harness.  At step $t$, the backend first
applies any event in $\Lambda$ whose trigger is satisfied, projects the latent state into the policy observation, and constructs the request payload.  
The model emits one typed action; the backend checks its preconditions against the complete latent state, applies its effect if executable, and returns the physical or symbolic event outcome.

The next request contains the 16 most recent inference actions and their outcomes.  
Interaction terminates when the model emits \texttt{stop} or reaches 100 steps.

The policy receives an action-skill catalog and must ground each skill's arguments in the current observation.
It is not provided with a list of grounded valid actions, thereby adhering to the open-world formulation.
The symbolic policy receives only visible nodes and relations, while the complete latent graph is retained internally by the backend only for transition execution and metric computation.

\paragraphb{System prompt.}
The agent uses the following system prompt:
\begin{quote}
\small\ttfamily
You are a policy for high-level embodied task evaluation.\\
Choose exactly one semantic action using the action catalog and the provided
observation.\\
Output strict JSON only.
\end{quote}

\paragraphb{Complete action catalog.}
Table~\ref{tab:appendix-action-catalog} expands every field of every action catalog
entry that can appear in a closed-loop evaluation request.  
Tabletop requests contain ten actions, while navigation-enabled indoor requests additionally contain \texttt{walk}.  

\paragraphb{Complete user payload.}
The user message contains a text block formed by serializing the following
object.  
The \texttt{action\_catalog} object contains the complete records in Table~\ref{tab:appendix-action-catalog} and all keys are shown explicitly below.
Figures~\ref{fig:appendix-payload}--\ref{fig:appendix-payload-output}
partition the serialized JSON format.

\begin{figure*}[t]
\centering
\begin{minipage}{0.96\textwidth}
\scriptsize
\begin{verbatim}
{
  "task": <high-level task title>,
  "task_type": <task-family identifier>,
  "settings": [<tabletop or indoor settings>],
  "success_criterion": <active symbolic goal g_t>,
  "step": <t>,
  "action_catalog": {
    "observe":    <complete catalog record>,
    "walk":       <complete record; indoor navigation only>,
    "open":       <complete catalog record>,
    "close":      <complete catalog record>,
    "press":      <complete catalog record>,
    "grab":       <complete catalog record>,
    "attach":     <complete catalog record>,
    "puton":      <complete catalog record>,
    "putin":      <complete catalog record>,
    "move_aside": <complete catalog record>,
    "stop":       <complete catalog record>
  },
  "action_constraints": [
    "Choose exactly one action using action_catalog and the provided
     observation.",
    "Ground object, target, and node_ids in the visual observation and task
     context.",
    "Do not assume that a candidate action list is available.",
    "Treat success_criterion as the authoritative task goal; task is only a
     high-level title.",
    "Treat recover as an internal transition, not as a second physical
     action.",
    "The current observation was captured after the previous task action and
     before the current action.",
    "Use the current observation to assess the most recent task action in
     recent_history; do not use that decision's prior recovery value as its
     outcome.",
    "Set recovery.required=true when the current observation shows that the
     most recent task action failed and the selected current action corrects
     that failure's consequences or retries the failed action.",
    "Set recovery.required=false when the current observation shows that it
     succeeded."
  ]
}
\end{verbatim}
\end{minipage}
\caption{Closed-loop user payload, part 1 of 3: task fields, the complete set
of action-catalog keys, and all symbolic action constraints.}
\label{fig:appendix-payload}
\end{figure*}

\begin{figure*}[t]
\centering
\begin{minipage}{0.96\textwidth}
\scriptsize
\begin{verbatim}
{
  "observation_timing": {
    "current_observation":
      "after the previous task action, before the selected current action",
    "recovery_assessment":
      "judge the previous task action from the current observation"
  },
  "recent_history": [
    {
      "step": <t-k>,
      "action": {
        "name": <previous semantic action>,
        "node_ids": [<ordered grounded identifiers>]
      },
      "event": {
        "status": <success or failure>,
        "failure_type": <typed failure or null>
      }
    }
    // the 16 most recent inference steps
  ],
  "current_observation": {
    "visible_nodes": [
      {
        "id": <node identifier>,
        "name": <node name>,
        "category": <node category>,
        "open": <boolean, when applicable>,
        "assembled": <boolean, when applicable>,
        "pressed": <boolean, when applicable>,
        "is_full": <boolean, when applicable>
      }
    ],
    "visible_edges": [
      {"from": <visible id>, "to": <visible id>, "relation": <relation>}
    ],
    "held_objects": [{"id": <held id>, "name": <held name>}],
    "robot": {
      "hands": <left/right hand state>,
      "navigation_enabled": <boolean, when applicable>,
      "current_room": <room id, when applicable>,
      "adjacent_rooms": [<room ids, when applicable>]
    }
  }
}
\end{verbatim}
\end{minipage}
\caption{Closed-loop user payload, part 2 of 3: observation timing, the
16-step inference-history window, and the partial visible-graph observation.}
\label{fig:appendix-payload-context}
\end{figure*}

\begin{figure*}[t]
\centering
\begin{minipage}{0.96\textwidth}
\scriptsize
\begin{verbatim}
{
  "output_schema": {
    "reason":
      "short rationale that assesses the previous action outcome from the
       current observation, then explains the selected current action",
    "recovery": {
      "required":
        "true when the current observation shows the previous task action
         failed and this action corrects its consequences or retries it",
      "failed_action":
        "previous task action name when required, otherwise null",
      "failed_node_ids":
        "node_ids of the previous failed action when required, otherwise [];
         order is ignored for attach and preserved for directional actions"
    },
    "action": {
      "name": "one semantic action name from action_catalog",
      "object": "observed object identifier when applicable",
      "target": "observed target identifier when applicable",
      "node_ids":
        "ordered observed object and target identifiers when applicable"
    }
  }
}
\end{verbatim}
\end{minipage}
\caption{Closed-loop user payload, part 3 of 3: the complete structured output
schema.}
\label{fig:appendix-payload-output}
\end{figure*}

For visual evaluation, the system prompt and all non-observation fields in the user prompt remain unchanged.  
The adapter replaces \texttt{current\_observation} with RGB camera references and appends the corresponding images to the user message.


\section{Metric Formulation}
\label{app:ssal}

\paragraphb{Success weighted by Symbolic Action Length.}
Let $S_r\in\{0,1\}$ indicate that rollout $r$ both satisfies the final goal
and terminates with a correct \texttt{stop}.  Let $L_{o,r}$ be the exact
symbolic oracle's shortest action-sequence length and let
\begin{equation}
    N_{s,r}=\sum_{t=1}^{T_r}\mathbf{1}[a_t\neq\texttt{stop}]
    \label{eq:non-stop-actions}
\end{equation}
be the model's number of non-stop action steps.  We define
\begin{equation}
    \operatorname{SSAL}_r
    =
    S_r
    \frac{L_{o,r}}{\max(L_{o,r},N_{s,r})}.
    \label{eq:ssal}
\end{equation}
Thus, unsuccessful rollouts receive zero, successful oracle-length rollouts
receive one, and additional non-stop actions reduce the score.  The maximum
in the denominator keeps the metric in $[0,1]$ if an observed execution is
shorter than the symbolic reference.

SR and the state-management metrics in Table~\ref{tab:models} average over the baseline and all six intervention conditions.  
SSAL in Table~\ref{tab:models} is averaged over the baseline condition only.

\section{Simulation-Backend Rollout Visualization}
\label{app:simulation-rollout}

\begin{figure*}[t]
\centering
\includegraphics[width=\textwidth,height=0.86\textheight,keepaspectratio]{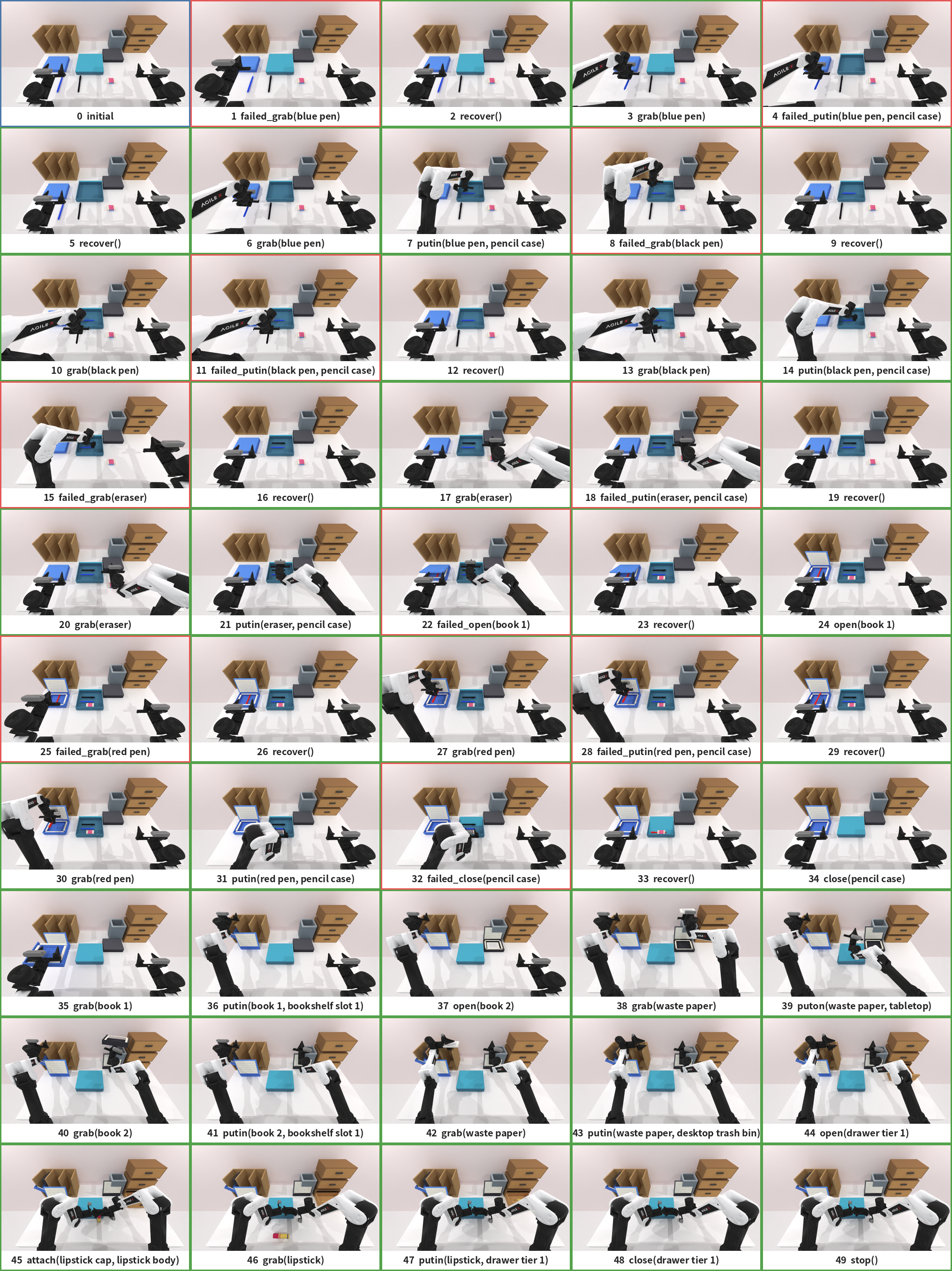}
\caption{All head-camera observations in a successful RoboTwin 2.0
backend-evaluation rollout (GPT-5.5). 
Blue marks the initial observation, red marks an injected failed action, and green marks a successfully executed transition.  
Each tile is labeled with its step index and issued action.}
\label{fig:appendix-robotwin-rollout}
\end{figure*}

\paragraphb{Successful physical rollout.}
Figure~\ref{fig:appendix-robotwin-rollout} visualizes every head-camera
observation from a successful 49-step RoboTwin backend-validation rollout in
assisted execution controller, including the initial frame.  
The trajectory contains injected failures and explicit recovery transitions before
completing all object placements at step 47, restoring the required closed state at step 48, and emitting \texttt{stop} at step 49.  

\paragraphb{Step-level model outputs.}
Table~\ref{tab:simulation-raw-responses} reports the issued action and the complete \texttt{raw\_response} stored for every action step in the same rollout.  

\section{Per-Scene Results and Difficulty Analysis}
\label{app:scene-results}

\paragraphb{Difficulty definition.}
We define empirical scene difficulty by mean SR over all 15 models and all seven conditions for that scene.  
NGP is reported alongside SR to distinguish complete failure from substantial partial progress.

\begin{table*}[t]
\centering
\begin{minipage}[t]{0.52\textwidth}
\centering
\scriptsize
\setlength{\tabcolsep}{2.6pt}
\begin{tabular}{cllrrrr}
\toprule
\textbf{Domain} & \textbf{Scene} & \textbf{Task family} & \textbf{SR} & \textbf{NGP} & $\boldsymbol{L_o}$ & $\boldsymbol{N_g}$ \\
\midrule
Tabletop & \texttt{Cosmetics-16} & Cosmetics storage & 4.8 & 46.6 & 30 & 20 \\
 & \texttt{Cosmetics-1} & Cosmetics storage & 4.8 & 52.8 & 28 & 20 \\
 & \texttt{Cosmetics-14} & Cosmetics storage & 5.7 & 38.5 & 30 & 20 \\
 & \texttt{Cosmetics-9} & Cosmetics storage & 8.6 & 40.1 & 30 & 20 \\
 & \texttt{Dining-8} & Dining-table organization & 9.5 & 39.9 & 19 & 9 \\
\addlinespace[2pt]
Indoor & \texttt{FoodPack-I4} & Food packing & 5.7 & 29.3 & 22 & 12 \\
 & \texttt{Fridge-I3} & Fridge-condiment loading & 6.7 & 39.9 & 23 & 14 \\
 & \texttt{Fridge-I1} & Fridge-condiment loading & 7.6 & 29.3 & 24 & 14 \\
 & \texttt{Fridge-I8} & Fridge-condiment loading & 7.6 & 40.4 & 24 & 14 \\
 & \texttt{Buffet-I7} & Buffet-tray sorting & 10.5 & 39.4 & 18 & 10 \\
\bottomrule
\end{tabular}
\caption{The five hardest scenes in each domain.  SR and NGP are percentages
averaged over all 15 models and seven conditions; $L_o$ and $N_g$ are the
symbolic oracle length and subgoal count.}
\label{tab:appendix-hardest-scenes}
\end{minipage}\hfill
\begin{minipage}[t]{0.45\textwidth}
\centering
\scriptsize
\setlength{\tabcolsep}{2.4pt}
\begin{tabular}{llrrrr}
\toprule
\textbf{Domain} & \textbf{Task family} & \textbf{Scenes} & \textbf{SR} & \textbf{NGP} & \textbf{RAR} \\
\midrule
Indoor & Buffet-tray sorting & 7 & 31.4 & 49.8 & 26.8 \\
Indoor & Food packing & 10 & 32.0 & 46.2 & 31.9 \\
Indoor & Fridge-condiment loading & 10 & 16.9 & 44.7 & 39.2 \\
Tabletop & Buffet-tray sorting & 7 & 41.1 & 59.4 & 33.2 \\
Tabletop & Cosmetics storage & 10 & 11.8 & 46.1 & 56.8 \\
Tabletop & Desk organization & 10 & 25.7 & 58.7 & 42.3 \\
Tabletop & Dining-table organization & 10 & 23.4 & 53.8 & 56.1 \\
Tabletop & Fridge-condiment loading & 10 & 25.0 & 52.5 & 34.2 \\
Tabletop & Toy organization & 10 & 29.0 & 64.1 & 30.4 \\
\bottomrule
\end{tabular}
\caption{Results aggregated by task family.  SR, NGP, and RAR are percentages
averaged over all scene--model pairs and seven conditions.}
\label{tab:appendix-family-results}
\end{minipage}
\vspace{-2ex}
\end{table*}

\paragraphb{Which scenes are difficult?}
Four of the five hardest tabletop scenes are cosmetics-storage instances. \texttt{Cosmetics-16} and \texttt{Cosmetics-1} each attain only 4.8\% SR, despite NGPs of 46.6\% and 52.8\%, respectively.  
This gap indicates that agents often make meaningful progress but fail to maintain and complete the full 20-unit goal.  
At the task-family level, cosmetics storage is the hardest tabletop family (11.8\% SR), followed by dining-table organization (23.4\% SR).  
Both also have high RARs, 56.8\% and 56.1\%, suggesting that repeated manipulation is a major failure mode.

For indoor tasks, \texttt{FoodPack-I4} is hardest at 5.7\% SR, followed by three fridge-condiment scenes with SR no higher than 7.6\%.  
Aggregated over scenes, fridge-condiment loading is the hardest indoor family at 16.9\% SR and has the highest indoor RAR at 39.2\%.  
Its difficulty is consistent with the need to coordinate room-level navigation, container opening and closing, and capacity-aware placement, which jointly involve both spatial and temporal dependencies.

\paragraphb{Complete per-scene results.}
Table~\ref{tab:per-scene-results} reports all 84 scenes and per-scene results averaged over all 15 models and seven conditions. 
Scenes are ordered from lower to higher SR within each domain.

\onecolumn
\vsize=\textheight
\begin{longtable}{@{}lllrrrrrrrrrr@{}}
\caption{Per-scene results averaged over all 15 models and seven conditions. Scenes are ordered from lower to higher SR within each domain.}
\label{tab:per-scene-results}\\
\toprule
\textbf{Domain} & \textbf{Scene} & \textbf{Family} & \textbf{SR} & \textbf{NGP} & \textbf{SSAL} & \textbf{AER} & \textbf{RAR} & \textbf{SEG} & \textbf{ADR} & \textbf{IRR} & $\boldsymbol{L_o}$ & $\boldsymbol{N_g}$ \\
\midrule
\endfirsthead
\multicolumn{13}{c}{\tablename~\thetable\ (continued)}\\
\toprule
\textbf{Domain} & \textbf{Scene} & \textbf{Family} & \textbf{SR} & \textbf{NGP} & \textbf{SSAL} & \textbf{AER} & \textbf{RAR} & \textbf{SEG} & \textbf{ADR} & \textbf{IRR} & $\boldsymbol{L_o}$ & $\boldsymbol{N_g}$ \\
\midrule
\endhead
\midrule
\multicolumn{13}{r}{Continued on next page}\\
\endfoot
\bottomrule
\endlastfoot
T & \texttt{Cosmetics-16} & Cosmetics & 4.8 & 46.6 & 0.034 & 61.6 & 54.8 & 41.5 & 7.6 & 4.4 & 30 & 20 \\
T & \texttt{Cosmetics-1} & Cosmetics & 4.8 & 52.8 & 0.000 & 64.8 & 48.0 & 42.7 & 5.8 & 5.6 & 28 & 20 \\
T & \texttt{Cosmetics-14} & Cosmetics & 5.7 & 38.5 & 0.060 & 62.3 & 64.4 & 36.1 & 6.1 & 5.0 & 30 & 20 \\
T & \texttt{Cosmetics-9} & Cosmetics & 8.6 & 40.1 & 0.000 & 62.8 & 63.2 & 36.0 & 6.7 & 10.7 & 30 & 20 \\
T & \texttt{Cosmetics-10} & Cosmetics & 9.5 & 39.0 & 0.053 & 63.4 & 60.3 & 42.2 & 8.4 & 10.0 & 30 & 20 \\
T & \texttt{Dining-8} & Dining & 9.5 & 39.9 & 0.057 & 59.6 & 72.3 & 23.4 & 7.4 & 16.3 & 19 & 9 \\
T & \texttt{Fridge-T9} & Fridge & 9.5 & 44.6 & 0.068 & 66.5 & 38.1 & 58.3 & 20.7 & 9.1 & 20 & 14 \\
T & \texttt{Cosmetics-13} & Cosmetics & 13.3 & 41.5 & 0.059 & 67.4 & 54.5 & 35.4 & 7.3 & 13.6 & 28 & 20 \\
T & \texttt{Cosmetics-12} & Cosmetics & 14.3 & 47.3 & 0.061 & 65.7 & 60.2 & 42.3 & 7.8 & 14.4 & 30 & 20 \\
T & \texttt{Toy-B2} & Toy & 14.3 & 59.8 & 0.172 & 69.2 & 36.2 & 51.8 & 5.6 & 14.3 & 31 & 16 \\
T & \texttt{Dining-7} & Dining & 14.3 & 63.1 & 0.089 & 59.8 & 51.9 & 13.0 & 2.5 & 16.4 & 18 & 9 \\
T & \texttt{Cosmetics-15} & Cosmetics & 15.2 & 48.3 & 0.067 & 66.2 & 58.6 & 41.6 & 7.7 & 15.6 & 30 & 20 \\
T & \texttt{Toy-A1} & Toy & 16.2 & 64.3 & 0.119 & 72.6 & 29.1 & 34.7 & 3.5 & 16.7 & 29 & 16 \\
T & \texttt{Dining-10} & Dining & 16.2 & 44.2 & 0.053 & 70.2 & 60.7 & 23.7 & 7.5 & 19.3 & 19 & 9 \\
T & \texttt{Desk-7} & Desk & 17.1 & 48.4 & 0.120 & 72.1 & 50.1 & 50.8 & 10.9 & 16.7 & 24 & 21 \\
T & \texttt{Toy-B1} & Toy & 17.1 & 61.8 & 0.239 & 75.9 & 25.0 & 27.8 & 2.9 & 14.7 & 29 & 16 \\
T & \texttt{Desk-4} & Desk & 18.1 & 58.2 & 0.102 & 71.4 & 44.3 & 46.3 & 11.0 & 18.9 & 24 & 21 \\
T & \texttt{Dining-9} & Dining & 19.0 & 49.0 & 0.176 & 56.3 & 60.4 & 42.5 & 9.5 & 19.0 & 19 & 9 \\
T & \texttt{Fridge-T2} & Fridge & 20.0 & 46.5 & 0.208 & 70.7 & 40.5 & 64.6 & 16.0 & 19.8 & 19 & 14 \\
T & \texttt{Cosmetics-4} & Cosmetics & 20.0 & 56.2 & 0.138 & 66.7 & 53.3 & 42.0 & 6.4 & 20.5 & 30 & 20 \\
T & \texttt{Desk-2} & Desk & 20.0 & 67.4 & 0.094 & 71.1 & 45.8 & 50.7 & 6.5 & 20.0 & 22 & 21 \\
T & \texttt{Toy-A5} & Toy & 21.0 & 61.3 & 0.192 & 76.8 & 23.9 & 36.8 & 3.6 & 21.1 & 31 & 16 \\
T & \texttt{Toy-A2} & Toy & 21.0 & 65.3 & 0.172 & 78.5 & 26.7 & 27.0 & 1.9 & 21.1 & 29 & 16 \\
T & \texttt{Fridge-T8} & Fridge & 21.9 & 47.8 & 0.230 & 72.5 & 34.3 & 64.6 & 20.4 & 21.3 & 21 & 14 \\
T & \texttt{Cosmetics-11} & Cosmetics & 21.9 & 50.3 & 0.204 & 67.8 & 51.0 & 40.6 & 6.9 & 21.3 & 29 & 20 \\
T & \texttt{Desk-11} & Desk & 21.9 & 55.8 & 0.159 & 72.2 & 47.0 & 52.2 & 10.6 & 22.2 & 24 & 21 \\
T & \texttt{Desk-15} & Desk & 21.9 & 65.5 & 0.243 & 75.5 & 32.1 & 49.6 & 10.5 & 21.3 & 24 & 21 \\
T & \texttt{Fridge-T1} & Fridge & 22.9 & 43.9 & 0.160 & 73.7 & 37.2 & 51.6 & 7.7 & 23.8 & 21 & 14 \\
T & \texttt{Dining-11} & Dining & 22.9 & 50.3 & 0.119 & 64.0 & 63.7 & 26.6 & 6.9 & 26.4 & 19 & 9 \\
T & \texttt{Fridge-T10} & Fridge & 22.9 & 54.1 & 0.124 & 65.7 & 37.4 & 69.9 & 14.8 & 24.9 & 18 & 14 \\
T & \texttt{Desk-8} & Desk & 23.8 & 44.4 & 0.230 & 73.1 & 46.2 & 51.4 & 21.5 & 23.1 & 25 & 21 \\
T & \texttt{Fridge-T3} & Fridge & 23.8 & 48.0 & 0.197 & 68.1 & 39.7 & 61.5 & 20.2 & 23.6 & 20 & 14 \\
T & \texttt{Fridge-T4} & Fridge & 25.7 & 52.2 & 0.262 & 72.9 & 36.1 & 60.1 & 10.6 & 27.0 & 20 & 14 \\
T & \texttt{Dining-5} & Dining & 25.7 & 52.6 & 0.205 & 76.2 & 51.1 & 23.5 & 5.9 & 26.8 & 19 & 9 \\
T & \texttt{Desk-13} & Desk & 25.7 & 56.9 & 0.227 & 73.6 & 44.7 & 53.4 & 12.1 & 24.0 & 24 & 21 \\
T & \texttt{Buffet-T4} & Buffet & 25.7 & 57.9 & 0.184 & 80.7 & 44.2 & 68.5 & 13.2 & 27.4 & 11 & 10 \\
T & \texttt{Desk-14} & Desk & 26.7 & 53.6 & 0.156 & 68.6 & 51.2 & 50.2 & 11.7 & 28.0 & 24 & 21 \\
T & \texttt{Fridge-T6} & Fridge & 26.7 & 58.1 & 0.288 & 76.9 & 24.7 & 71.8 & 16.0 & 29.0 & 18 & 14 \\
T & \texttt{Fridge-T7} & Fridge & 26.7 & 63.4 & 0.267 & 80.8 & 22.3 & 67.8 & 20.5 & 27.1 & 18 & 14 \\
T & \texttt{Dining-3} & Dining & 26.7 & 74.2 & 0.246 & 74.5 & 38.2 & 23.0 & 4.1 & 27.6 & 18 & 9 \\
T & \texttt{Desk-12} & Desk & 27.6 & 57.9 & 0.227 & 68.7 & 37.8 & 49.5 & 10.9 & 26.7 & 24 & 21 \\
T & \texttt{Dining-6} & Dining & 28.6 & 48.4 & 0.149 & 71.6 & 60.6 & 36.2 & 8.5 & 33.3 & 19 & 9 \\
T & \texttt{Buffet-T6} & Buffet & 29.5 & 41.7 & 0.158 & 81.4 & 55.2 & 49.0 & 7.7 & 30.4 & 11 & 10 \\
T & \texttt{Buffet-T7} & Buffet & 29.5 & 53.9 & 0.229 & 82.9 & 42.8 & 64.6 & 8.8 & 28.9 & 15 & 10 \\
T & \texttt{Dining-12} & Dining & 35.2 & 58.1 & 0.252 & 69.6 & 47.2 & 43.6 & 12.1 & 36.9 & 19 & 9 \\
T & \texttt{Toy-B5} & Toy & 35.2 & 63.0 & 0.326 & 69.6 & 35.8 & 67.6 & 8.5 & 33.8 & 31 & 16 \\
T & \texttt{Dining-4} & Dining & 36.2 & 58.4 & 0.233 & 68.7 & 55.1 & 33.8 & 7.1 & 37.8 & 19 & 9 \\
T & \texttt{Buffet-T2} & Buffet & 37.1 & 42.5 & 0.328 & 81.0 & 36.8 & 39.0 & 8.2 & 40.7 & 10 & 10 \\
T & \texttt{Toy-B3} & Toy & 37.1 & 68.7 & 0.347 & 76.1 & 27.7 & 28.4 & 2.2 & 34.7 & 27 & 16 \\
T & \texttt{Toy-B4} & Toy & 40.0 & 64.5 & 0.339 & 74.6 & 39.4 & 28.6 & 3.3 & 39.3 & 27 & 16 \\
T & \texttt{Toy-A3} & Toy & 43.8 & 65.6 & 0.497 & 76.5 & 30.8 & 11.9 & 0.9 & 42.2 & 25 & 16 \\
T & \texttt{Toy-A4} & Toy & 44.8 & 66.6 & 0.332 & 77.8 & 28.9 & 25.1 & 2.3 & 45.6 & 27 & 16 \\
T & \texttt{Buffet-T3} & Buffet & 49.5 & 68.3 & 0.559 & 86.2 & 23.6 & 84.4 & 9.0 & 48.7 & 10 & 10 \\
T & \texttt{Fridge-T5} & Fridge & 50.5 & 66.7 & 0.501 & 69.5 & 31.6 & 24.6 & 0.3 & 49.8 & 18 & 14 \\
T & \texttt{Desk-1} & Desk & 54.3 & 78.5 & 0.456 & 78.6 & 24.2 & 47.3 & 6.3 & 54.4 & 20 & 21 \\
T & \texttt{Buffet-T5} & Buffet & 58.1 & 73.3 & 0.542 & 85.4 & 17.1 & 84.4 & 12.8 & 56.7 & 12 & 10 \\
T & \texttt{Buffet-T1} & Buffet & 58.1 & 78.3 & 0.650 & 90.7 & 13.1 & 90.3 & 16.4 & 57.8 & 14 & 10 \\
\addlinespace[3pt]
I & \texttt{FoodPack-I4} & Food pack & 5.7 & 29.3 & 0.031 & 74.8 & 37.6 & 24.3 & 4.0 & 5.8 & 22 & 12 \\
I & \texttt{Fridge-I3} & Fridge & 6.7 & 39.9 & 0.099 & 67.9 & 41.3 & 34.4 & 7.6 & 7.5 & 23 & 14 \\
I & \texttt{Fridge-I1} & Fridge & 7.6 & 29.3 & 0.041 & 72.6 & 42.6 & 24.9 & 2.7 & 8.0 & 24 & 14 \\
I & \texttt{Fridge-I8} & Fridge & 7.6 & 40.4 & 0.000 & 71.7 & 38.4 & 36.5 & 8.8 & 8.0 & 24 & 14 \\
I & \texttt{Buffet-I7} & Buffet & 10.5 & 39.4 & 0.046 & 72.8 & 35.1 & 22.5 & 4.6 & 12.0 & 18 & 10 \\
I & \texttt{Fridge-I9} & Fridge & 10.5 & 43.0 & 0.093 & 73.0 & 38.2 & 34.7 & 8.4 & 10.0 & 23 & 14 \\
I & \texttt{Fridge-I2} & Fridge & 11.4 & 41.1 & 0.030 & 70.7 & 37.4 & 36.2 & 8.2 & 11.4 & 22 & 14 \\
I & \texttt{Buffet-I4} & Buffet & 14.3 & 52.4 & 0.028 & 74.1 & 27.4 & 21.2 & 8.1 & 15.6 & 14 & 10 \\
I & \texttt{Fridge-I4} & Fridge & 17.1 & 40.8 & 0.112 & 69.5 & 46.7 & 30.2 & 4.5 & 20.0 & 23 & 14 \\
I & \texttt{Fridge-I7} & Fridge & 18.1 & 47.9 & 0.139 & 66.5 & 36.8 & 31.6 & 7.6 & 17.1 & 21 & 14 \\
I & \texttt{Buffet-I6} & Buffet & 23.8 & 31.2 & 0.095 & 72.7 & 33.6 & 14.6 & 5.6 & 25.1 & 14 & 10 \\
I & \texttt{Fridge-I10} & Fridge & 23.8 & 56.9 & 0.189 & 68.7 & 34.4 & 38.4 & 7.2 & 25.8 & 21 & 14 \\
I & \texttt{FoodPack-I1} & Food pack & 25.7 & 41.7 & 0.197 & 73.2 & 31.6 & 26.8 & 2.6 & 26.7 & 24 & 12 \\
I & \texttt{Fridge-I6} & Fridge & 25.7 & 49.3 & 0.170 & 63.8 & 43.6 & 34.6 & 6.5 & 29.6 & 21 & 14 \\
I & \texttt{Buffet-I2} & Buffet & 28.6 & 35.9 & 0.207 & 75.9 & 30.6 & 7.1 & 2.9 & 27.1 & 12 & 10 \\
I & \texttt{FoodPack-I8} & Food pack & 28.6 & 42.6 & 0.243 & 73.3 & 32.5 & 27.0 & 2.8 & 27.0 & 24 & 12 \\
I & \texttt{FoodPack-I9} & Food pack & 32.4 & 46.2 & 0.133 & 74.9 & 31.0 & 26.4 & 3.3 & 33.1 & 21 & 12 \\
I & \texttt{FoodPack-I10} & Food pack & 33.3 & 40.0 & 0.201 & 73.6 & 29.4 & 22.9 & 3.3 & 33.6 & 21 & 12 \\
I & \texttt{FoodPack-I7} & Food pack & 33.3 & 54.0 & 0.272 & 75.8 & 31.8 & 28.3 & 4.1 & 32.2 & 21 & 12 \\
I & \texttt{FoodPack-I5} & Food pack & 39.0 & 47.1 & 0.254 & 72.5 & 32.5 & 30.7 & 3.0 & 38.2 & 21 & 12 \\
I & \texttt{FoodPack-I3} & Food pack & 39.0 & 53.3 & 0.298 & 73.6 & 30.5 & 32.1 & 6.6 & 38.9 & 21 & 12 \\
I & \texttt{FoodPack-I2} & Food pack & 40.0 & 57.4 & 0.270 & 75.5 & 32.8 & 29.9 & 4.1 & 41.3 & 21 & 12 \\
I & \texttt{Fridge-I5} & Fridge & 40.0 & 58.9 & 0.205 & 72.2 & 32.1 & 19.9 & 0.4 & 38.7 & 21 & 14 \\
I & \texttt{FoodPack-I6} & Food pack & 42.9 & 50.5 & 0.298 & 73.8 & 29.6 & 32.4 & 3.2 & 40.0 & 23 & 12 \\
I & \texttt{Buffet-I3} & Buffet & 44.8 & 59.0 & 0.275 & 76.5 & 16.3 & 17.8 & 4.0 & 48.1 & 13 & 10 \\
I & \texttt{Buffet-I5} & Buffet & 48.6 & 62.4 & 0.389 & 77.8 & 22.4 & 24.4 & 6.1 & 50.5 & 15 & 10 \\
I & \texttt{Buffet-I1} & Buffet & 49.5 & 68.4 & 0.340 & 77.2 & 22.3 & 33.2 & 8.2 & 49.6 & 17 & 10 \\
\end{longtable}

\begingroup
\setlength{\tabcolsep}{3.5pt}
\setlength{\LTleft}{0pt}
\setlength{\LTright}{0pt}
\footnotesize
\onecolumn
\begin{longtable}{@{}r >{\raggedright\arraybackslash}p{0.19\textwidth} >{\raggedright\arraybackslash}p{0.68\textwidth}@{}}
\caption{Step-level issued actions and model (GPT-5.5) \texttt{raw\_response} strings for a successful rollout.}
\label{tab:simulation-raw-responses}\\
\toprule
\textbf{Step} & \textbf{Issued action} & \textbf{\texttt{raw\_response}} \\
\midrule
\endfirsthead
\multicolumn{3}{c}{\tablename\ \thetable\ (continued)} \\
\toprule
\textbf{Step} & \textbf{Issued action} & \textbf{\texttt{raw\_response}} \\
\midrule
\endhead
\midrule
\multicolumn{3}{r}{Continued on next page} \\
\endfoot
\bottomrule
\endlastfoot
1 & {\scriptsize\ttfamily failed\_grab(blue pen)} & {\scriptsize\ttfamily \{\newline{}  "reason": "The task requires placing three pens and an eraser into the pencil case. Currently, the robot's hands are empty. The valid actions include grabbing the 'blue pen', 'black pen', or 'eraser'. I will start by grabbing one of the items to be placed in the pencil case. Let's grab the blue pen first.",\newline{}  "action": \{\newline{}    "name": "grab",\newline{}    "object": "blue pen",\newline{}    "target": null,\newline{}    "node\_ids": [\newline{}      "blue pen"\newline{}    ]\newline{}  \}\newline{}\}} \\
\addlinespace[2pt]
2 & {\scriptsize\ttfamily recover()} & {\scriptsize\ttfamily \{\newline{}  "reason": "The previous action 'grab' on 'blue pen' failed due to an injected failure. According to the action constraints, after an injected failure, valid\_actions contains 'recover' and 'stop'. To continue the task, I must choose 'recover' to retry the previous action.",\newline{}  "action": \{\newline{}    "name": "recover",\newline{}    "object": null,\newline{}    "target": null,\newline{}    "node\_ids": []\newline{}  \}\newline{}\}} \\
\addlinespace[2pt]
3 & {\scriptsize\ttfamily grab(blue pen)} & {\scriptsize\ttfamily \{\newline{}  "reason": "The previous 'grab' action on 'blue pen' failed due to an injected failure, but 'recover' was successfully executed in the last step. According to the constraints, after a successful recovery, the valid\_actions will contain only the original failed action so it can be retried. The current valid\_actions list contains exactly one option: grabbing the 'blue pen'. Therefore, I must retry this action.",\newline{}  "action": \{\newline{}    "name": "grab",\newline{}    "object": "blue pen",\newline{}    "target": null,\newline{}    "node\_ids": [\newline{}      "blue pen"\newline{}    ]\newline{}  \}\newline{}\}} \\
\addlinespace[2pt]
4 & {\scriptsize\ttfamily failed\_putin(blue pen, pencil case)} & {\scriptsize\ttfamily \{\newline{}  "reason": "The robot is currently holding the blue pen. The task requires placing the blue pen into the pencil box. The pencil box is visible and open. Therefore, the next logical step is to put the held blue pen into the pencil box.",\newline{}  "action": \{\newline{}    "name": "putin",\newline{}    "object": "blue pen",\newline{}    "target": "pencil case",\newline{}    "node\_ids": [\newline{}      "blue pen",\newline{}      "pencil case"\newline{}    ]\newline{}  \}\newline{}\}} \\
\addlinespace[2pt]
5 & {\scriptsize\ttfamily recover()} & {\scriptsize\ttfamily \{\newline{}  "reason": "The previous action 'putin' failed due to an injected failure. According to the action constraints, after an injected failure, valid\_actions contains 'recover' and 'stop'. To continue the task and retry the failed action, I must choose 'recover'.",\newline{}  "action": \{\newline{}    "name": "recover",\newline{}    "object": null,\newline{}    "target": null,\newline{}    "node\_ids": []\newline{}  \}\newline{}\}} \\
\addlinespace[2pt]
6 & {\scriptsize\ttfamily grab(blue pen)} & {\scriptsize\ttfamily ""} \emph{(empty in source trajectory)} \\
\addlinespace[2pt]
7 & {\scriptsize\ttfamily putin(blue pen, pencil case)} & {\scriptsize\ttfamily \{\newline{}  "reason": "The previous 'putin' action failed due to an injected failure, but 'recover' was successfully executed in the last step. According to the constraints, after a successful recovery, the valid\_actions will contain only the original failed action so it can be retried. The current valid\_actions list contains exactly one option: putting the 'blue pen' into the 'pencil case'. Therefore, I must retry this action.",\newline{}  "action": \{\newline{}    "name": "putin",\newline{}    "object": "blue pen",\newline{}    "target": "pencil case",\newline{}    "node\_ids": [\newline{}      "blue pen",\newline{}      "pencil case"\newline{}    ]\newline{}  \}\newline{}\}} \\
\addlinespace[2pt]
8 & {\scriptsize\ttfamily failed\_grab(black pen)} & {\scriptsize\ttfamily \{\newline{}  "reason": "The blue pen has been successfully placed in the pencil case. The next step is to place the remaining items (black pen, red pen, eraser) into the pencil case. The black pen is visible on the desk and grabbable. I will grab the black pen next.",\newline{}  "action": \{\newline{}    "name": "grab",\newline{}    "object": "black pen",\newline{}    "target": null,\newline{}    "node\_ids": [\newline{}      "black pen"\newline{}    ]\newline{}  \}\newline{}\}} \\
\addlinespace[2pt]
9 & {\scriptsize\ttfamily recover()} & {\scriptsize\ttfamily \{\newline{}  "reason": "The previous action 'grab' on 'black pen' failed due to an injected failure. According to the action constraints, after an injected failure, valid\_actions contains 'recover' and 'stop'. To continue the task and retry the failed action, I must choose 'recover'.",\newline{}  "action": \{\newline{}    "name": "recover",\newline{}    "object": null,\newline{}    "target": null,\newline{}    "node\_ids": []\newline{}  \}\newline{}\}} \\
\addlinespace[2pt]
10 & {\scriptsize\ttfamily grab(black pen)} & {\scriptsize\ttfamily \{\newline{}  "reason": "The previous 'grab' action on 'black pen' failed due to an injected failure, but 'recover' was successfully executed in the last step. According to the constraints, after a successful recovery, the valid\_actions will contain only the original failed action so it can be retried. The current valid\_actions list contains exactly one option: grabbing the 'black pen'. Therefore, I must retry this action.",\newline{}  "action": \{\newline{}    "name": "grab",\newline{}    "object": "black pen",\newline{}    "target": null,\newline{}    "node\_ids": [\newline{}      "black pen"\newline{}    ]\newline{}  \}\newline{}\}} \\
\addlinespace[2pt]
11 & {\scriptsize\ttfamily failed\_putin(black pen, pencil case)} & {\scriptsize\ttfamily \{\newline{}  "reason": "The robot is currently holding the black pen. The task requires placing the black pen into the pencil box. The pencil box is visible and open. Therefore, the next logical step is to put the held black pen into the pencil box.",\newline{}  "action": \{\newline{}    "name": "putin",\newline{}    "object": "black pen",\newline{}    "target": "pencil case",\newline{}    "node\_ids": [\newline{}      "black pen",\newline{}      "pencil case"\newline{}    ]\newline{}  \}\newline{}\}} \\
\addlinespace[2pt]
12 & {\scriptsize\ttfamily recover()} & {\scriptsize\ttfamily \{\newline{}  "reason": "The previous action 'putin' on 'black pen' and 'pencil case' failed due to an injected failure. According to the action constraints, after an injected failure, valid\_actions contains 'recover' and 'stop'. To continue the task and retry the failed action, I must choose 'recover'.",\newline{}  "action": \{\newline{}    "name": "recover",\newline{}    "object": null,\newline{}    "target": null,\newline{}    "node\_ids": []\newline{}  \}\newline{}\}} \\
\addlinespace[2pt]
13 & {\scriptsize\ttfamily grab(black pen)} & {\scriptsize\ttfamily ""} \emph{(empty in source trajectory)} \\
\addlinespace[2pt]
14 & {\scriptsize\ttfamily putin(black pen, pencil case)} & {\scriptsize\ttfamily \{\newline{}  "reason": "The previous 'putin' action for the black pen failed due to an injected failure, but 'recover' was successfully executed in the last step. According to the constraints, after a successful recovery, the valid\_actions will contain only the original failed action so it can be retried. The current valid\_actions list contains exactly one option: putting the 'black pen' into the 'pencil case'. Therefore, I must retry this action.",\newline{}  "action": \{\newline{}    "name": "putin",\newline{}    "object": "black pen",\newline{}    "target": "pencil case",\newline{}    "node\_ids": [\newline{}      "black pen",\newline{}      "pencil case"\newline{}    ]\newline{}  \}\newline{}\}} \\
\addlinespace[2pt]
15 & {\scriptsize\ttfamily failed\_grab(eraser)} & {\scriptsize\ttfamily \{\newline{}  "reason": "The blue and black pens are already in the pencil case. The red pen is not visible, but the eraser is visible on the desk. The task requires placing the eraser into the pencil case. Since the robot's hands are empty, the next logical step is to grab the eraser.",\newline{}  "action": \{\newline{}    "name": "grab",\newline{}    "object": "eraser",\newline{}    "target": null,\newline{}    "node\_ids": [\newline{}      "eraser"\newline{}    ]\newline{}  \}\newline{}\}} \\
\addlinespace[2pt]
16 & {\scriptsize\ttfamily recover()} & {\scriptsize\ttfamily \{\newline{}  "reason": "The previous action 'grab' on 'eraser' failed due to an injected failure. According to the action constraints, after an injected failure, valid\_actions contains 'recover' and 'stop'. To continue the task and retry the failed action, I must choose 'recover'.",\newline{}  "action": \{\newline{}    "name": "recover",\newline{}    "object": null,\newline{}    "target": null,\newline{}    "node\_ids": []\newline{}  \}\newline{}\}} \\
\addlinespace[2pt]
17 & {\scriptsize\ttfamily grab(eraser)} & {\scriptsize\ttfamily \{\newline{}  "reason": "The previous action 'grab' on 'eraser' failed due to an injected failure, but 'recover' was successfully executed in the last step. According to the constraints, after a successful recovery, the valid\_actions will contain only the original failed action so it can be retried. The current valid\_actions list contains exactly one option: grabbing the 'eraser'. Therefore, I must retry this action.",\newline{}  "action": \{\newline{}    "name": "grab",\newline{}    "object": "eraser",\newline{}    "target": null,\newline{}    "node\_ids": [\newline{}      "eraser"\newline{}    ]\newline{}  \}\newline{}\}} \\
\addlinespace[2pt]
18 & {\scriptsize\ttfamily failed\_putin(eraser, pencil case)} & {\scriptsize\ttfamily \{\newline{}  "reason": "The robot is currently holding the eraser. The task requires placing the eraser into the pencil box. The pencil box is visible and open. Therefore, the next logical step is to put the held eraser into the pencil box.",\newline{}  "action": \{\newline{}    "name": "putin",\newline{}    "object": "eraser",\newline{}    "target": "pencil case",\newline{}    "node\_ids": [\newline{}      "eraser",\newline{}      "pencil case"\newline{}    ]\newline{}  \}\newline{}\}} \\
\addlinespace[2pt]
19 & {\scriptsize\ttfamily recover()} & {\scriptsize\ttfamily \{\newline{}  "reason": "The previous action 'putin' on 'eraser' and 'pencil case' failed due to an injected failure. According to the action constraints, after an injected failure, valid\_actions contains 'recover' and 'stop'. To continue the task and retry the failed action, I must choose 'recover'.",\newline{}  "action": \{\newline{}    "name": "recover",\newline{}    "object": null,\newline{}    "target": null,\newline{}    "node\_ids": []\newline{}  \}\newline{}\}} \\
\addlinespace[2pt]
20 & {\scriptsize\ttfamily grab(eraser)} & {\scriptsize\ttfamily ""} \emph{(empty in source trajectory)} \\
\addlinespace[2pt]
21 & {\scriptsize\ttfamily putin(eraser, pencil case)} & {\scriptsize\ttfamily \{\newline{}  "reason": "The previous 'putin' action for the eraser failed due to an injected failure, but 'recover' was successfully executed in the last step. According to the constraints, after a successful recovery, the valid\_actions will contain only the original failed action so it can be retried. The current valid\_actions list contains exactly one option: putting the 'eraser' into the 'pencil case'. Therefore, I must retry this action.",\newline{}  "action": \{\newline{}    "name": "putin",\newline{}    "object": "eraser",\newline{}    "target": "pencil case",\newline{}    "node\_ids": [\newline{}      "eraser",\newline{}      "pencil case"\newline{}    ]\newline{}  \}\newline{}\}} \\
\addlinespace[2pt]
22 & {\scriptsize\ttfamily failed\_open(book 1)} & {\scriptsize\ttfamily \{\newline{}  "reason": "The blue pen, black pen, and eraser are already inside the pencil case. The red pen is not visible in the current observation, but the task requires closing the pencil box after placing items. Since the pencil box is currently open and contains the required items (assuming the red pen is either already inside or not visible/required for this step's immediate action given the valid actions), and 'close' is a valid action for the pencil box, I should close it. However, looking at the success criteria, the red pen must be inside. The red pen is not visible. But wait, the valid actions do not include searching for the red pen. Let's re-read the valid actions. The valid actions are: open books/drawers, grab eraser/blue/black pens, close pencil box, stop. The eraser, blue pen, and black pen are already inside the pencil box according to the observation (location: INSIDE target: pencil case). The red pen is missing from the visible nodes. This might imply it's hidden or the task description implies it's there. However, I cannot act on invisible objects. The only progress-making action available that aligns with the goal of 'closing the pencil box' is to close it. But first, I need to ensure all items are in. The red pen is not visible. Is it possible the red pen is inside one of the books or drawers? The books are closed. The drawers are closed. I should probably check inside the books or drawers. But 'open' is a valid action for books and drawers. Let's look at the constraints. 'Book itself cannot be loaded into other items'. This doesn't prevent opening. Let's try opening a book to see if the red pen is there, or just close the pencil box if we assume the red pen is handled. Actually, looking at the history, we haven't looked for the red pen. But 'look' is not in valid\_actions. 'inspect' is not in valid\_actions. 'open' is. Let's open 'book 1' to see if the red pen is inside.",\newline{}  "action": \{\newline{}    "name": "open",\newline{}    "object": "book 1",\newline{}    "target": null,\newline{}    "node\_ids": [\newline{}      "book 1"\newline{}    ]\newline{}  \}\newline{}\}} \\
\addlinespace[2pt]
23 & {\scriptsize\ttfamily recover()} & {\scriptsize\ttfamily \{\newline{}  "reason": "The previous action 'open' on 'book 1' failed due to an injected failure. According to the action constraints, after an injected failure, valid\_actions contains 'recover' and 'stop'. To continue the task and retry the failed action, I must choose 'recover'.",\newline{}  "action": \{\newline{}    "name": "recover",\newline{}    "object": null,\newline{}    "target": null,\newline{}    "node\_ids": []\newline{}  \}\newline{}\}} \\
\addlinespace[2pt]
24 & {\scriptsize\ttfamily open(book 1)} & {\scriptsize\ttfamily \{\newline{}  "reason": "The previous action 'open' on 'book 1' failed due to an injected failure, but 'recover' was successfully executed in the last step. According to the constraints, after a successful recovery, the valid\_actions will contain only the original failed action so it can be retried. The current valid\_actions list contains exactly one option: opening 'book 1'. Therefore, I must retry this action.",\newline{}  "action": \{\newline{}    "name": "open",\newline{}    "object": "book 1",\newline{}    "target": null,\newline{}    "node\_ids": [\newline{}      "book 1"\newline{}    ]\newline{}  \}\newline{}\}} \\
\addlinespace[2pt]
25 & {\scriptsize\ttfamily failed\_grab(red pen)} & {\scriptsize\ttfamily \{\newline{}  "reason": "The red pen was just revealed inside book\_1. The task requires placing the red pen into the pencil case. The pencil case is currently open. The robot's hands are empty. Therefore, the next logical step is to grab the red pen.",\newline{}  "action": \{\newline{}    "name": "grab",\newline{}    "object": "red pen",\newline{}    "target": null,\newline{}    "node\_ids": [\newline{}      "red pen"\newline{}    ]\newline{}  \}\newline{}\}} \\
\addlinespace[2pt]
26 & {\scriptsize\ttfamily recover()} & {\scriptsize\ttfamily \{\newline{}  "reason": "The previous action 'grab' on 'red pen' failed due to an injected failure. According to the action constraints, after an injected failure, valid\_actions contains 'recover' and 'stop'. To continue the task and retry the failed action, I must choose 'recover'.",\newline{}  "action": \{\newline{}    "name": "recover",\newline{}    "object": null,\newline{}    "target": null,\newline{}    "node\_ids": []\newline{}  \}\newline{}\}} \\
\addlinespace[2pt]
27 & {\scriptsize\ttfamily grab(red pen)} & {\scriptsize\ttfamily \{\newline{}  "reason": "The previous action 'grab' on 'red pen' failed due to an injected failure, but 'recover' was successfully executed in the last step. According to the constraints, after a successful recovery, the valid\_actions will contain only the original failed action so it can be retried. The current valid\_actions list contains exactly one option: grabbing the 'red pen'. Therefore, I must retry this action.",\newline{}  "action": \{\newline{}    "name": "grab",\newline{}    "object": "red pen",\newline{}    "target": null,\newline{}    "node\_ids": [\newline{}      "red pen"\newline{}    ]\newline{}  \}\newline{}\}} \\
\addlinespace[2pt]
28 & {\scriptsize\ttfamily failed\_putin(red pen, pencil case)} & {\scriptsize\ttfamily \{\newline{}  "reason": "The robot is currently holding the red pen. The task requires placing the red pen into the pencil box. The pencil box is visible and open. Therefore, the next logical step is to put the held red pen into the pencil box.",\newline{}  "action": \{\newline{}    "name": "putin",\newline{}    "object": "red pen",\newline{}    "target": "pencil case",\newline{}    "node\_ids": [\newline{}      "red pen",\newline{}      "pencil case"\newline{}    ]\newline{}  \}\newline{}\}} \\
\addlinespace[2pt]
29 & {\scriptsize\ttfamily recover()} & {\scriptsize\ttfamily \{\newline{}  "reason": "The previous action 'putin' on 'red pen' and 'pencil case' failed due to an injected failure. According to the action constraints, after an injected failure, valid\_actions contains 'recover' and 'stop'. To continue the task and retry the failed action, I must choose 'recover'.",\newline{}  "action": \{\newline{}    "name": "recover",\newline{}    "object": null,\newline{}    "target": null,\newline{}    "node\_ids": []\newline{}  \}\newline{}\}} \\
\addlinespace[2pt]
30 & {\scriptsize\ttfamily grab(red pen)} & {\scriptsize\ttfamily ""} \emph{(empty in source trajectory)} \\
\addlinespace[2pt]
31 & {\scriptsize\ttfamily putin(red pen, pencil case)} & {\scriptsize\ttfamily \{\newline{}  "reason": "The previous 'putin' action for the red pen failed due to an injected failure, but 'recover' was successfully executed in the last step. According to the constraints, after a successful recovery, the valid\_actions will contain only the original failed action so it can be retried. The current valid\_actions list contains exactly one option: putting the 'red pen' into the 'pencil case'. Therefore, I must retry this action.",\newline{}  "action": \{\newline{}    "name": "putin",\newline{}    "object": "red pen",\newline{}    "target": "pencil case",\newline{}    "node\_ids": [\newline{}      "red pen",\newline{}      "pencil case"\newline{}    ]\newline{}  \}\newline{}\}} \\
\addlinespace[2pt]
32 & {\scriptsize\ttfamily failed\_close(pencil case)} & {\scriptsize\ttfamily \{\newline{}  "reason": "All three pens and the eraser are now inside the pencil box. The next step in the task is to close the pencil box. The 'close' action for 'pencil case' is available in valid\_actions.",\newline{}  "action": \{\newline{}    "name": "close",\newline{}    "object": "pencil case",\newline{}    "target": null,\newline{}    "node\_ids": [\newline{}      "pencil case"\newline{}    ]\newline{}  \}\newline{}\}} \\
\addlinespace[2pt]
33 & {\scriptsize\ttfamily recover()} & {\scriptsize\ttfamily \{\newline{}  "reason": "The previous action 'close' on 'pencil case' failed due to an injected failure. According to the action constraints, after an injected failure, valid\_actions contains 'recover' and 'stop'. To continue the task and retry the failed action, I must choose 'recover'.",\newline{}  "action": \{\newline{}    "name": "recover",\newline{}    "object": null,\newline{}    "target": null,\newline{}    "node\_ids": []\newline{}  \}\newline{}\}} \\
\addlinespace[2pt]
34 & {\scriptsize\ttfamily close(pencil case)} & {\scriptsize\ttfamily \{\newline{}  "reason": "The previous action 'close' on 'pencil case' failed due to an injected failure, but 'recover' was successfully executed in the last step. According to the constraints, after a successful recovery, the valid\_actions will contain only the original failed action so it can be retried. The current valid\_actions list contains exactly one option: closing the 'pencil case'. Therefore, I must retry this action.",\newline{}  "action": \{\newline{}    "name": "close",\newline{}    "object": "pencil case",\newline{}    "target": null,\newline{}    "node\_ids": [\newline{}      "pencil case"\newline{}    ]\newline{}  \}\newline{}\}} \\
\addlinespace[2pt]
35 & {\scriptsize\ttfamily grab(book 1)} & {\scriptsize\ttfamily \{\newline{}  "reason": "The pencil box has been successfully closed. The next part of the task is to place the two books into the first shelf. Currently, 'book 1' is open and on the desk, and 'book 2' is closed and on the desk. To place them in the shelf, I first need to ensure they are ready to be moved. 'book 1' is currently open; while it can be grabbed, closing it might be a prerequisite or good practice before moving, but more importantly, I need to grab the books to put them in the shelf. Looking at valid\_actions, I can 'grab' 'book 1'. I will start by grabbing the first book.",\newline{}  "action": \{\newline{}    "name": "grab",\newline{}    "object": "book 1",\newline{}    "target": null,\newline{}    "node\_ids": [\newline{}      "book 1"\newline{}    ]\newline{}  \}\newline{}\}} \\
\addlinespace[2pt]
36 & {\scriptsize\ttfamily putin(book 1, bookshelf slot 1)} & {\scriptsize\ttfamily \{\newline{}  "reason": "The robot is currently holding 'book 1'. The task requires placing both books into 'bookshelf slot 1'. Since 'book 1' is already held and 'bookshelf slot 1' is a valid target container, the next logical step is to place the held book into the shelf.",\newline{}  "action": \{\newline{}    "name": "putin",\newline{}    "object": "book 1",\newline{}    "target": "bookshelf slot 1",\newline{}    "node\_ids": [\newline{}      "book 1",\newline{}      "bookshelf slot 1"\newline{}    ]\newline{}  \}\newline{}\}} \\
\addlinespace[2pt]
37 & {\scriptsize\ttfamily open(book 2)} & {\scriptsize\ttfamily \{\newline{}  "reason": "Book 1 has been successfully placed in the first shelf. The next step is to place Book 2 into the first shelf. Book 2 is currently on the desk and needs to be grabbed first. The valid actions include grabbing 'book 1' (which is already in the shelf, so likely not the target) or other actions. Wait, looking at valid\_actions, I see 'grab' for 'book 1'. But 'book 1' is already in the shelf. I need to grab 'book 2'. However, 'book 2' is not in the valid\_actions list for 'grab'. Let me re-examine valid\_actions. The valid actions are: close 'book 1', open 'book 2', open drawers, open pencil box, grab 'book 1', stop. This is strange. 'book 2' is visible and reachable. Why is 'grab' 'book 2' not valid? Perhaps because 'book 2' is closed and acts as a container? No, it's grabbable. Let's look closer. Ah, 'book 1' is currently OPEN. The valid action 'close' 'book 1' is available. Maybe I should close it? No, the task is to put books in the shelf. 'book 1' is already there. I need to move 'book 2'. If 'grab' 'book 2' is not available, maybe I need to inspect it or something else? Or maybe I made a mistake and 'book 1' refers to the object ID which might be reusable? No. Let's look at the state. 'book 1' is INSIDE 'bookshelf slot 1'. 'book 2' is ON 'tabletop'. I need to get 'book 2' into 'bookshelf slot 1'. The only grab action available is for 'book 1'. This seems like a trap or I am missing something. Wait, if 'book 1' is in the shelf, grabbing it would take it out. That's counter-productive. Is it possible that 'book 2' is not grabbable because it's considered a container that needs to be opened? No, it has GRABBABLE property. Let's re-read the valid actions carefully. \textbackslash{}nValid actions:\textbackslash{}n1. close 'book 1'\textbackslash{}n2. open 'book 2'\textbackslash{}n3. open drawers...\textbackslash{}n4. open pencil box\textbackslash{}n5. grab 'book 1'\textbackslash{}n6. stop\textbackslash{}n\textbackslash{}nIf I cannot grab 'book 2', I cannot complete the task. However, often in these environments, if an object is a container, you might need to ensure it's closed or open depending on context. But 'putin' requires holding. \textbackslash{}nLet's consider that maybe I need to 'open' 'book 2' to make it manipulable or check inside? The hint says 'may\_reveal\_hidden'. There might be another item inside, or maybe opening it changes its state to allow grabbing? Or perhaps the 'grab' action for 'book 2' is missing due to some constraint I don't see (e.g. it's considered 'full' or something?). \textbackslash{}nActually, looking at previous steps, when 'red pen' was hidden, I had to open 'book 1' to reveal it. Maybe 'book 2' also contains something required? The success criteria doesn't mention anything inside book 2. \textbackslash{}nHowever, I MUST choose from valid actions. I cannot invent 'grab' 'book 2'. \textbackslash{}nBetween 'close' 'book 1' and 'open' 'book 2': \textbackslash{}nClosing 'book 1' might be good practice before leaving it, but it's already in the shelf. \textbackslash{}nOpening 'book 2' might reveal something or just change its state. \textbackslash{}nLet's try opening 'book 2' as it is the only action involving the second book, which is the remaining item to be processed.",\newline{}  "action": \{\newline{}    "name": "open",\newline{}    "object": "book 2",\newline{}    "target": null,\newline{}    "node\_ids": [\newline{}      "book 2"\newline{}    ]\newline{}  \}\newline{}\}} \\
\addlinespace[2pt]
38 & {\scriptsize\ttfamily grab(waste paper)} & {\scriptsize\ttfamily \{\newline{}  "reason": "The books are now in the shelf. The waste paper was revealed inside book\_2 and is now visible on the desk. The task requires putting the waste paper into the desktop trash can. The valid actions include grabbing the waste paper. I will grab it first.",\newline{}  "action": \{\newline{}    "name": "grab",\newline{}    "object": "waste paper",\newline{}    "target": null,\newline{}    "node\_ids": [\newline{}      "waste paper"\newline{}    ]\newline{}  \}\newline{}\}} \\
\addlinespace[2pt]
39 & {\scriptsize\ttfamily puton(waste paper, tabletop)} & {\scriptsize\ttfamily ""} \emph{(empty in source trajectory)} \\
\addlinespace[2pt]
40 & {\scriptsize\ttfamily grab(book 2)} & {\scriptsize\ttfamily \{\newline{}  "reason": "The previous step opened 'book 2', revealing 'waste paper'. The current valid actions include grabbing 'book 2'. Although the ultimate goal is to place 'book 2' in the shelf, it must be grabbed first. Since 'grab' for 'book 2' is a valid action and necessary to move it, I will select it.",\newline{}  "action": \{\newline{}    "name": "grab",\newline{}    "object": "book 2",\newline{}    "target": null,\newline{}    "node\_ids": [\newline{}      "book 2"\newline{}    ]\newline{}  \}\newline{}\}} \\
\addlinespace[2pt]
41 & {\scriptsize\ttfamily putin(book 2, bookshelf slot 1)} & {\scriptsize\ttfamily \{\newline{}  "reason": "The robot is currently holding 'book 2'. The task requires placing both books into 'bookshelf slot 1'. 'book 1' is already in the shelf. The next logical step is to place the held 'book 2' into 'bookshelf slot 1'. This action is available in valid\_actions.",\newline{}  "action": \{\newline{}    "name": "putin",\newline{}    "object": "book 2",\newline{}    "target": "bookshelf slot 1",\newline{}    "node\_ids": [\newline{}      "book 2",\newline{}      "bookshelf slot 1"\newline{}    ]\newline{}  \}\newline{}\}} \\
\addlinespace[2pt]
42 & {\scriptsize\ttfamily grab(waste paper)} & {\scriptsize\ttfamily ""} \emph{(empty in source trajectory)} \\
\addlinespace[2pt]
43 & {\scriptsize\ttfamily putin(waste paper, desktop trash bin)} & {\scriptsize\ttfamily \{\newline{}  "reason": "The robot is currently holding the waste paper. The task requires placing the waste paper into the desktop trash can ('desktop trash bin'). This action is available in valid\_actions and aligns with the success criterion.",\newline{}  "action": \{\newline{}    "name": "putin",\newline{}    "object": "waste paper",\newline{}    "target": "desktop trash bin",\newline{}    "node\_ids": [\newline{}      "waste paper",\newline{}      "desktop trash bin"\newline{}    ]\newline{}  \}\newline{}\}} \\
\addlinespace[2pt]
44 & {\scriptsize\ttfamily open(drawer tier 1)} & {\scriptsize\ttfamily \{\newline{}  "reason": "Most task goals are met: pens/eraser in pencil box (closed), books in shelf, waste paper in trash. However, the success criterion includes 'ASSEMBLED' for 'lipstick' and 'INSIDE' 'lipstick' in 'drawer tier 1'. These objects are not currently visible or handled. The valid actions include opening drawers, which may reveal the lipstick. Opening 'drawer tier 1' is a logical next step to find the missing item.",\newline{}  "action": \{\newline{}    "name": "open",\newline{}    "object": "drawer tier 1",\newline{}    "target": null,\newline{}    "node\_ids": [\newline{}      "drawer tier 1"\newline{}    ]\newline{}  \}\newline{}\}} \\
\addlinespace[2pt]
45 & {\scriptsize\ttfamily attach(lipstick cap, lipstick body)} & {\scriptsize\ttfamily \{\newline{}  "reason": "The lipstick parts (cap and body) have been revealed in the first drawer. The success criterion requires the lipstick to be assembled and placed in the first drawer. The valid actions include 'attach' for 'lipstick cap' and 'lipstick body'. This is the necessary step to assemble the lipstick.",\newline{}  "action": \{\newline{}    "name": "attach",\newline{}    "object": "lipstick cap",\newline{}    "target": "lipstick body",\newline{}    "node\_ids": [\newline{}      "lipstick cap",\newline{}      "lipstick body"\newline{}    ]\newline{}  \}\newline{}\}} \\
\addlinespace[2pt]
46 & {\scriptsize\ttfamily grab(lipstick)} & {\scriptsize\ttfamily \{\newline{}  "reason": "The lipstick has been assembled. The success criterion requires the lipstick to be inside the first drawer. The first drawer is currently open. The valid actions include grabbing the lipstick. I need to grab it first before putting it in the drawer.",\newline{}  "action": \{\newline{}    "name": "grab",\newline{}    "object": "lipstick",\newline{}    "target": null,\newline{}    "node\_ids": [\newline{}      "lipstick"\newline{}    ]\newline{}  \}\newline{}\}} \\
\addlinespace[2pt]
47 & {\scriptsize\ttfamily putin(lipstick, drawer tier 1)} & {\scriptsize\ttfamily \{\newline{}  "reason": "The robot is currently holding the assembled lipstick. The success criterion requires the lipstick to be inside the first drawer ('drawer tier 1'). The first drawer is currently open and visible. Therefore, the next logical step is to place the held lipstick into the first drawer using the 'putin' action.",\newline{}  "action": \{\newline{}    "name": "putin",\newline{}    "object": "lipstick",\newline{}    "target": "drawer tier 1",\newline{}    "node\_ids": [\newline{}      "lipstick",\newline{}      "drawer tier 1"\newline{}    ]\newline{}  \}\newline{}\}} \\
\addlinespace[2pt]
48 & {\scriptsize\ttfamily close(drawer tier 1)} & {\scriptsize\ttfamily ""} \emph{(empty in source trajectory)} \\
\addlinespace[2pt]
49 & {\scriptsize\ttfamily stop()} & {\scriptsize\ttfamily \{\newline{}  "reason": "All success criteria have been met: pens and eraser are in the closed pencil box, books are in the first shelf, waste paper is in the trash, and the assembled lipstick is in the first drawer. The task is complete.",\newline{}  "action": \{\newline{}    "name": "stop",\newline{}    "object": null,\newline{}    "target": null,\newline{}    "node\_ids": []\newline{}  \}\newline{}\}} \\
\addlinespace[2pt]
\end{longtable}
\endgroup

\end{document}